%% file: main.tex
\documentclass[sigconf, 9pt, nonacm]{acmart} 

\usepackage{amsmath}
\usepackage{tabularx}
\usepackage{array}
\usepackage{balance}
\usepackage{subcaption}
\usepackage{xspace}
\usepackage{multirow,siunitx}
\usepackage{lipsum}
\usepackage{enumitem}

\usepackage{graphicx}
\usepackage{booktabs}
\usepackage{subcaption}
\usepackage{cuted}

\newcolumntype{Y}{>{\centering\arraybackslash}X}

\newcommand{\sys}{mmAnomaly\xspace}
\newcommand{\parlabel}[1]{\vspace{0.5em}\noindent\textbf{#1}}

\newcommand\note[1]{\textcolor{black}{#1}}

\newcommand\bnote[1]{\textcolor{black}{#1}}

\begin{document}

\title{\sys: Leveraging Visual Context for Robust Anomaly Detection in the Non-Visual World with mmWave Radar}

\author{Tarik Reza Toha}
\orcid{0000-0002-6529-3487}
\affiliation{
  \institution{University of North Carolina at Chapel Hill}
  \city{Chapel Hill}
  \state{North Carolina}
  \country{USA}
}
\email{ttoha12@cs.unc.edu}

\author{Shao-Jung (Louie) Lu}
\orcid{0009-0007-8530-9283}
\affiliation{
  \institution{University of North Carolina at Chapel Hill}
  \city{Chapel Hill}
  \state{North Carolina}
  \country{USA}
}
\email{louielu@cs.unc.edu}

\author{Mahathir Monjur}
\orcid{0000-0002-2488-9911}
\affiliation{
  \institution{University of North Carolina at Chapel Hill}
  \city{Chapel Hill}
  \state{North Carolina}
  \country{USA}
}
\email{mahathir@cs.unc.edu}

\author{Shahriar Nirjon}
\orcid{0000-0003-1443-1146}
\affiliation{
  \institution{University of North Carolina at Chapel Hill}
  \city{Chapel Hill}
  \state{North Carolina}
  \country{USA}
}
\email{nirjon@cs.unc.edu}

\renewcommand{\shortauthors}{Toha et al.}

\input{tex/00_abstract}




\maketitle

\input{tex/01_introduction}

\input{tex/03_empirical}

\input{tex/04_method}
\input{tex/05_evalutaion}

\input{tex/06_applications}

\input{tex/07_discussion}
\input{tex/08_literature}
\input{tex/09_conclusion}

\balance
\bibliographystyle{ACM-Reference-Format}
\bibliography{bib}

\end{document}

%% file: tex/00_abstract.tex
\begin{abstract}
mmWave radar enables human sensing in non-visual scenarios—e.g., through clothing or certain types of walls—where traditional cameras fail due to occlusion or privacy limitations. However, robust anomaly detection with mmWave remains challenging, as signal reflections are influenced by material properties, clutter, and multipath interference, producing complex, non-Gaussian distortions. Existing methods lack contextual awareness and misclassify benign signal variations as anomalies. We present \sys, a multi-modal anomaly detection framework that combines mmWave radar with RGBD input to incorporate visual context. Our system extracts semantic cues—such as scene geometry and material properties—using a fast ResNet-based classifier, and uses a conditional latent diffusion model to synthesize the expected mmWave spectrum for the given visual context. A dual-input comparison module then identifies spatial deviations between real and generated spectra to localize anomalies. We evaluate \sys on two multi-modal datasets across three applications: concealed weapon localization, through-wall intruder localization, and through-wall fall localization. The system achieves up to 94\% F1 score \bnote{and sub-meter localization error,} demonstrating robust generalization across clothing, occlusions, and cluttered environments. These results establish \sys as an accurate and interpretable framework for context-aware anomaly detection in mmWave sensing.
\end{abstract}

%% file: tex/01_introduction.tex
\section{Introduction}
\label{sec:intro}

\parlabel{mmWave Anomaly Detection Applications.}  
mmWave anomaly detection has several critical applications, particularly in security \cite{gui2025enhanced, chen2024imageless, woodford2024metasight, zhang2021mmeye}, healthcare \cite{he2024radio, hao2024mmwave, tang2024bsense, zhang2023pivimo}, and industrial monitoring \cite{murakami2024development, sun2024integrated, gu2024mmbox}, where traditional sensors like cameras, infrared, and acoustic sensors fall short. In security screening, mmWave can penetrate clothing and detect concealed weapons or contraband that optical cameras fail to recognize \cite{gui2025enhanced, chen2024imageless}. In healthcare, mmWave enables non-contact physiological monitoring, such as detecting abnormal respiration patterns indicative of medical conditions like sleep apnea \cite{he2024radio, hao2024mmwave}. In industrial settings, mmWave can identify structural anomalies \cite{murakami2024development}, material defects \cite{sun2024integrated}, or equipment malfunctions \cite{gu2024mmbox} in harsh environments where visibility is limited. These applications highlight the unique advantages of mmWave sensing in scenarios where other modalities alone are ineffective.

\begin{figure}[!t]
\centering
\begin{subfigure}[c]{0.23\textwidth}
\centering
\includegraphics[width=\textwidth]{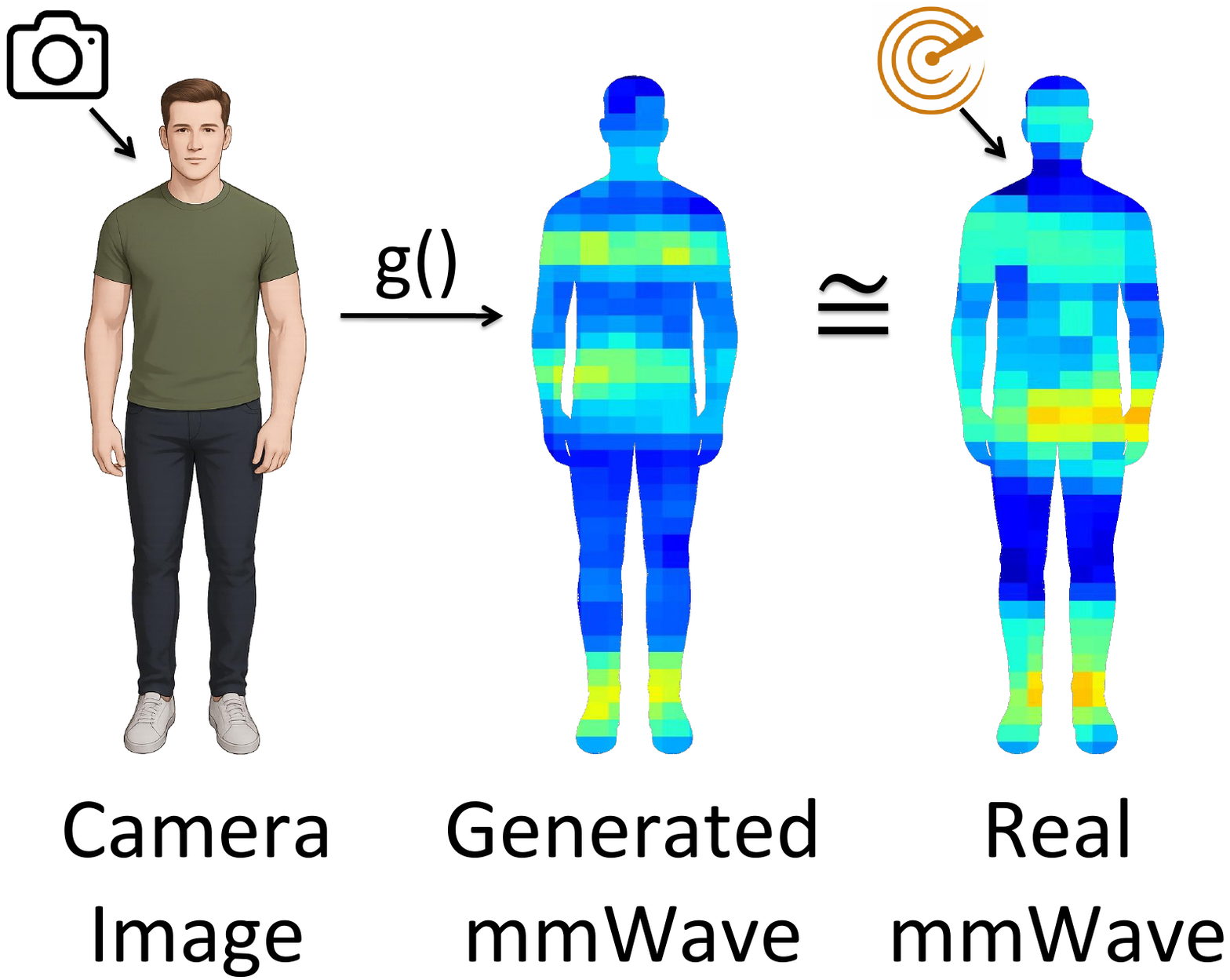}
\caption{without concealed weapon}
\end{subfigure}
\hfil
\begin{subfigure}[c]{0.23\textwidth}
\centering
\includegraphics[width=\textwidth]{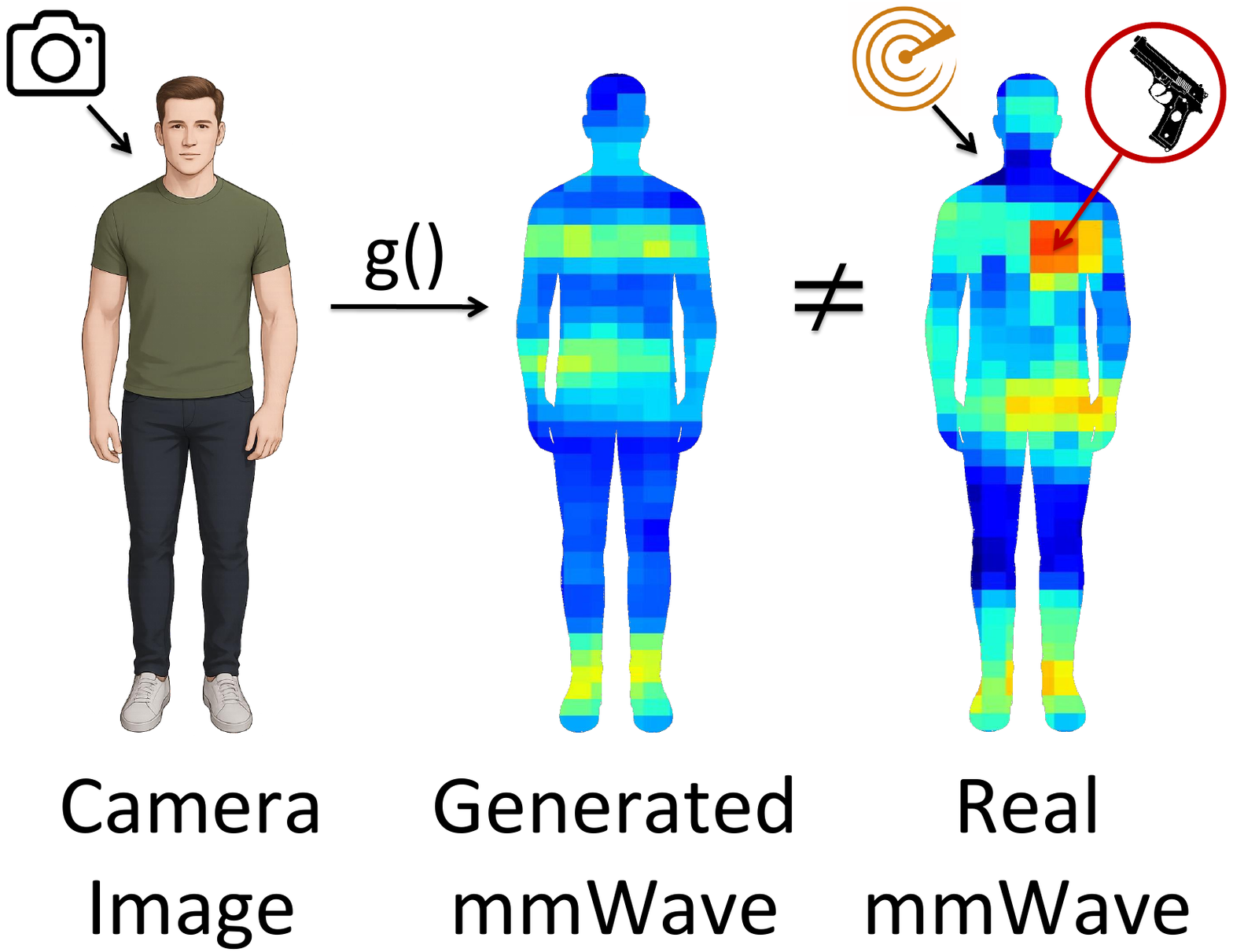}
\caption{with concealed weapon}
\end{subfigure}
\vspace{-2mm}
\caption{\sys detects and localizes anomaly (e.g., concealed weapons) by comparing real mmWave data with mmWave signals generated from RGBD images, using the fact that visual appearance stays the same while mmWave reflections change when hidden items are present.}
\label{fig:teaser}
\vspace{-2mm}
\end{figure}

\parlabel{Challenges in mmWave Anomaly Detection.}
The challenges of mmWave anomaly detection stem from the unique properties of mmWave signals, which differ fundamentally from other sensing modalities. Unlike visual or infrared data, mmWave signals exhibit strong sensitivity to environmental conditions, such as surface reflectivity, material properties, and multipath effects. Additionally, mmWave signals do not conform to the typical noise models assumed in traditional anomaly-detection algorithms. Instead, they are subject to complex propagation phenomena, including constructive and destructive interference, non-line-of-sight reflections, and diffraction, which evolve with scene dynamics and introduce structured yet non-stationary distortions. Moreover, anomalies in mmWave spectra rarely appear as isolated spikes or sparse outliers; they are often embedded within subtle waveform deviations or spatially dependent fluctuations that mirror benign environmental variation. These factors create a pronounced mismatch between the physics-driven nature of mmWave signals and the statistical assumptions of conventional models, making it difficult to distinguish true anomalies from context-induced variability without leveraging additional environmental cues.

\parlabel{Limitations of Existing Anomaly Detection Methods.}
Existing anomaly detection methods for mmWave radar fall into two broad categories, each with significant limitations. The first includes vision-inspired models such as GLAD~\cite{yao2024glad}, GLASS~\cite{chen2024unified}, and GeneralAD~\cite{strater2024generalad}, which are designed for structured signals like images or time series. When retrofitted for mmWave data, these models fail to capture subtle waveform distortions and structured interference patterns, often misclassifying benign variations as anomalies due to their reliance on simplistic outlier-based assumptions. The second category includes RF-specific models like SiFall~\cite{ji2023sifall} and CGAN~\cite{toma2020deep}, which are better aligned with wireless signals but still fall short for mmWave. These models assume simplified noise characteristics and/or overlook critical context such as scene geometry, material properties, and multipath effects. As a result, they cannot reliably distinguish true anomalies from context-driven variations, highlighting the need for multimodal approaches that incorporate visual cues.

\parlabel{Leveraging Visual Context.} While mmWave signals can detect anomalies hidden from visual sensors—such as concealed weapons, intruders behind walls, or falls behind obstructions—relying exclusively on radar data limits anomaly detection performance. mmWave reflections are highly sensitive to environmental and scene-specific factors like surface materials, furniture placement, and dynamic clutter, which can introduce significant variability and ambiguity. Although visual sensors cannot directly observe these hidden anomalies, they provide valuable context about visible environmental structures and scene layouts. By leveraging visual context from a co-located RGBD camera, anomaly detection algorithms can better calibrate their expectations of normal mmWave reflections given the observed materials and spatial arrangements. We conduct a thorough empirical study to understand the effect of incorporating visual context and observe that visual cues consistently enhance detection accuracy and robustness compared to mmWave-only models. This demonstrates that visual guidance plays a critical role in improving the reliability of mmWave anomaly detection systems.

\parlabel{We Propose \sys: A Cross-Modal Generative Approach.}
To operationalize visual context for mmWave anomaly detection, we introduce \emph{\sys}, a multi-modal system that bridges low-level radar signals with high-level scene understanding through cross-modal generation. \bnote{In this paper, we define an \emph{anomaly} as a rare mmWave-observable event that cannot be observed by the visible RGBD scene, focusing on two broad classes: (i) unusual physical body movements (e.g., abrupt falls or struggling motions), and (ii) unexpected metallic objects on or near the body (e.g., concealed weapons). Conceptually, \sys instantiates a new sensing paradigm: instead of relying on radar-only statistical priors, it uses a generative model to synthesize the \emph{expected} mmWave response from an anchor sensor (the camera). It treats deviations from this prediction as evidence of an anomaly.} 

To our knowledge, \sys is the first to synthesize expected mmWave signals directly from RGBD inputs of a co-located camera, enabling anomaly detection in scenarios where visual sensors or mmWave alone are insufficient. Trained solely on regular (anomaly-free) data, \sys learns how visual semantics—such as material properties, human pose, and environmental layout—shape radar reflections. At runtime, \sys compares the predicted mmWave response with the actual measurement; the resulting residual localizes anomalies in the non-visual domain, such as concealed weapons, intruders behind walls, or falls behind obstructions. We implement this idea by integrating RGBD perception modules with a conditional generative backbone and a residual-based anomaly localizer. By grounding radar interpretation in visual context, \sys reduces false positives, enhances robustness to environmental variability, and improves generalization across settings. Figure~\ref{fig:teaser} illustrates this principle: even when the RGBD view appears normal (e.g., a person walking naturally), \sys exploits subtle residuals between observed and predicted mmWave reflections to reveal and localize hidden threats such as concealed weapons.

\parlabel{Experimental Evaluation and Results.}  
We conduct an extensive evaluation of \sys on two multi-modal datasets spanning three representative applications: concealed object detection, through-wall intrusion detection, and fall detection—each posing distinct challenges for mmWave sensing. 

\begin{enumerate}[leftmargin=6mm]
\item For concealed object detection, we use a self-collected dataset where metallic items are hidden beneath various clothing types. \sys achieves an F1 score of approximately 94\% and a mean localization error of 16\,mm, outperforming CVAE~\cite{anand2025vae}, CGAN~\cite{toma2020deep}, and MMW-Carry~\cite{gao2024mmwcarry} across fleece, leather, and snow jackets under diverse environmental conditions. 

\item For through-wall intrusion detection, we evaluate \sys on a custom dataset involving human presence behind different wall types (e.g., Styrofoam, gator board, and particle board) with added reflectors. \sys attains an F1 score of $\sim$92\% and maintains low localization error ($\sim$0.7\,m), demonstrating strong generalization in cluttered and occluded environments. 

\item In fall detection, also using the custom dataset, \sys combines pose-aware classification with temporal entropy to distinguish fall events from routine activity, achieving $\sim$96\% F1 and $\sim$0.5\,m localization error, even when subjects are occluded or outside the camera view. 
\end{enumerate}

Overall, \sys consistently outperforms baseline methods in both detection and localization accuracy, validating the effectiveness of its multi-modal, context-aware design across varied and challenging scenarios.

%% file: tex/03_empirical.tex
\section{Empirical Study}
\label{sec:empirical}

We conduct an empirical study on the problem of mmWave-based anomaly detection with a two-fold objective: (1) to assess the performance of state-of-the-art anomaly detection models, and (2) to investigate the extent to which incorporating visual cues (e.g., clothing type and environmental settings) enhances detection performance on mmWave data.


\begin{figure}[!t]
\centering
\begin{subfigure}[c]{0.23\textwidth}
\centering
\includegraphics[width=\textwidth]{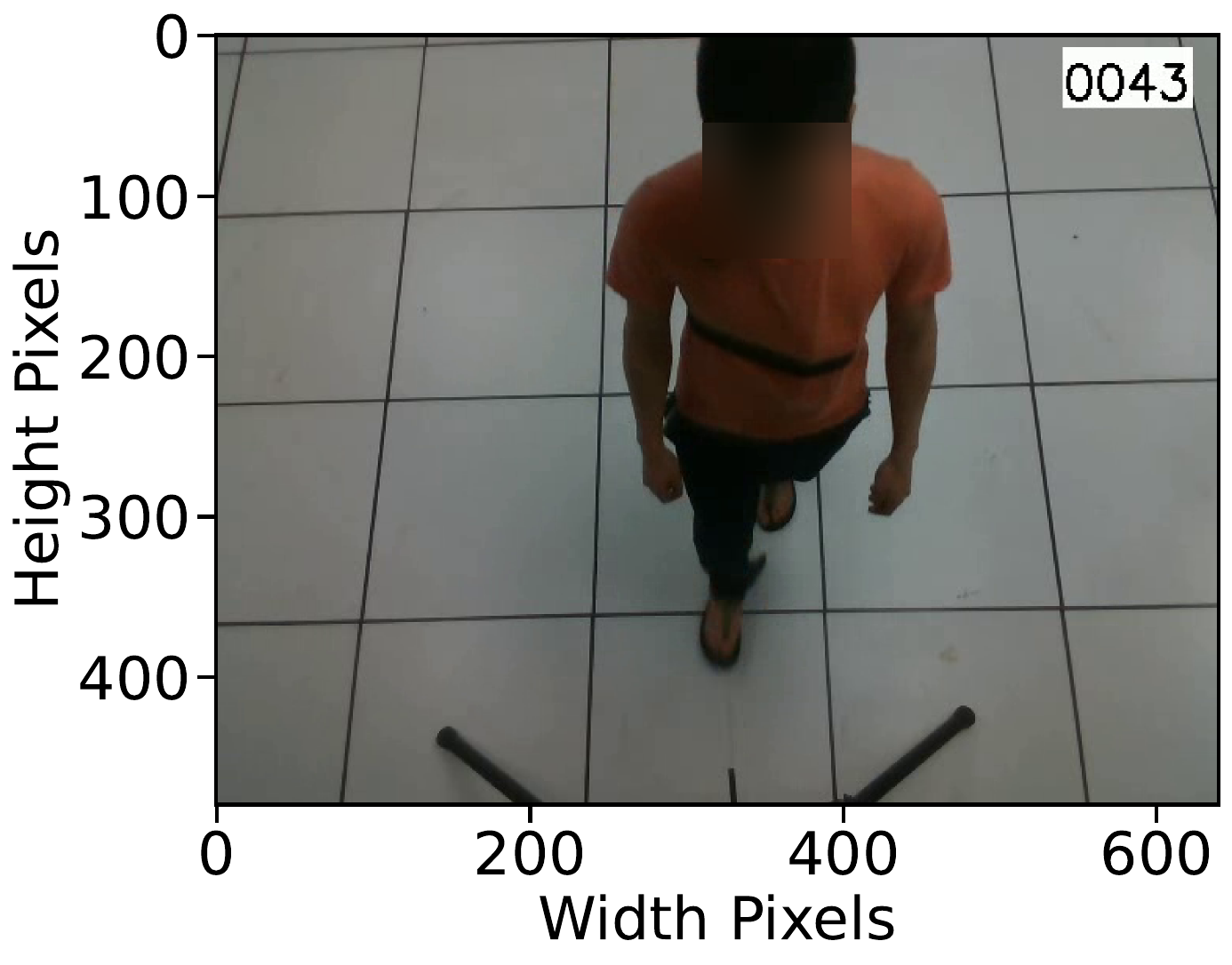}
\caption{RGB frame}
\end{subfigure}
\hfil
\begin{subfigure}[c]{0.23\textwidth}
\centering
\includegraphics[width=\textwidth]{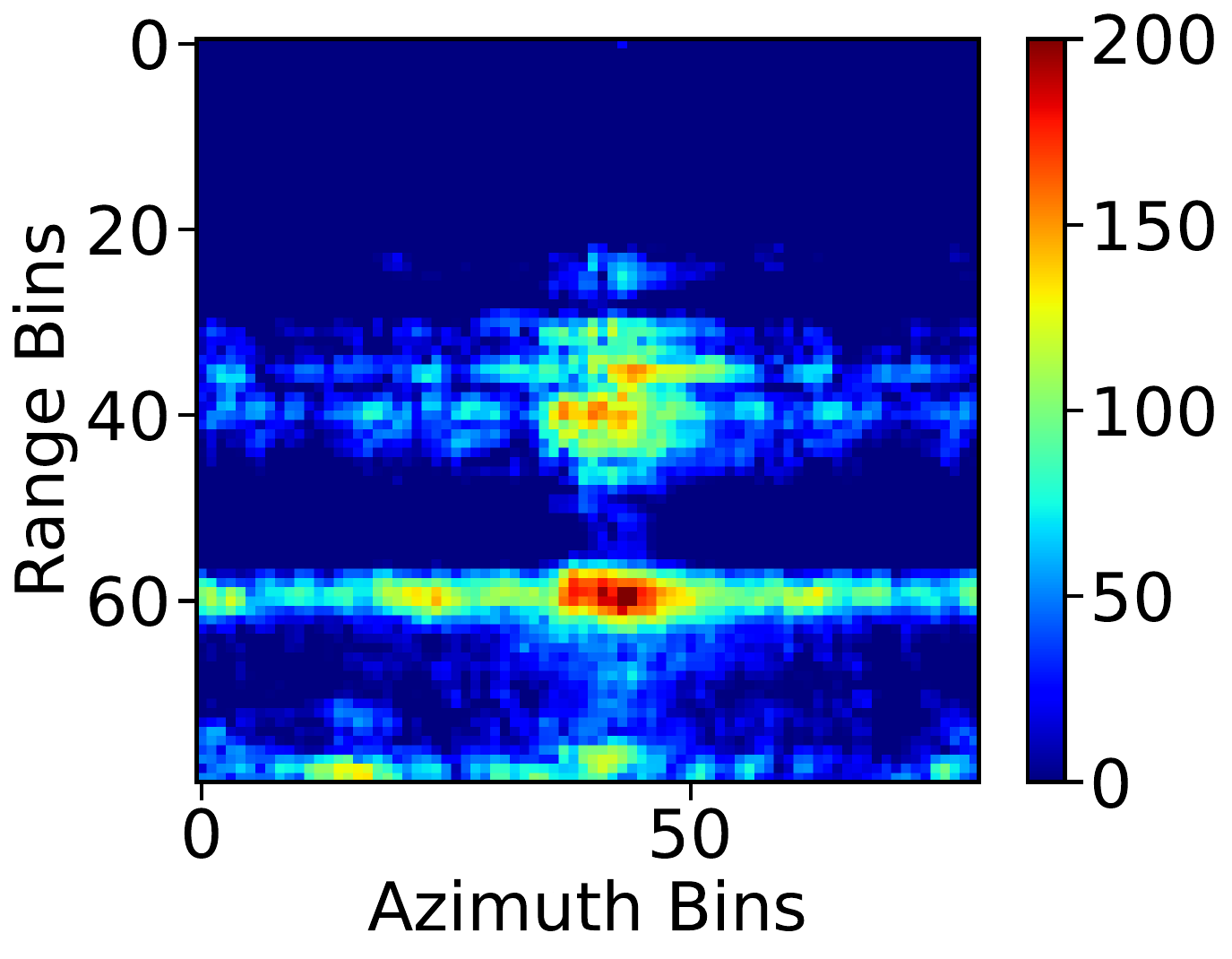}
\caption{Range-Azimuth spectrum}
\end{subfigure}
\caption{Example mmWave radar scene showing a person with a concealed weapon in the left pocket.}
\label{fig:motiv_scene}
\vspace{-3mm}
\end{figure}

\begin{figure}[!t]
\centering
\begin{subfigure}[c]{0.23\textwidth}
\centering
\includegraphics[width=\textwidth]{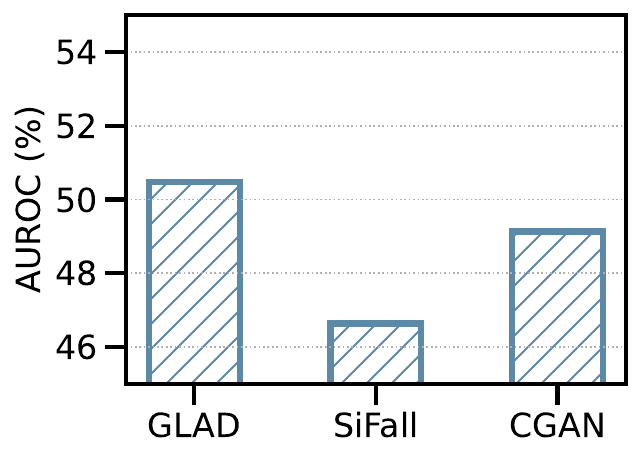}
\caption{AUROC ($\uparrow$ better)}
\end{subfigure}
\hfil
\begin{subfigure}[c]{0.23\textwidth}
\centering
\includegraphics[width=\textwidth]{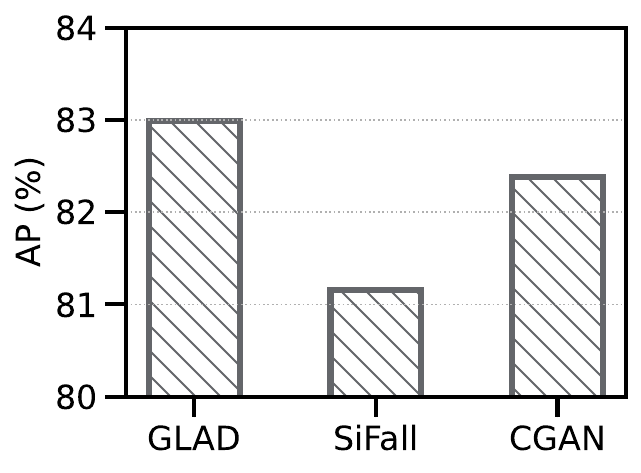}
\caption{Average Precision ($\uparrow$ better)}
\end{subfigure}
\caption{Anomaly detection performance of baseline models using only mmWave radar inputs.}
\label{fig:soa_results}
\vspace{-3mm}
\end{figure}

\subsection{Experimental Setup}

\parlabel{Dataset.} We use a self-collected dataset, which contains approximately 1,200 mmWave radar recordings of seven participants walking past a sensor mounted with a \ang{20} downward tilt. Each 4-second sequence (60 frames at 15 fps) is labeled as \textit{normal} (no object) or \textit{anomalous}, where anomalies involve one of three metallic weapon-like objects concealed at one of eight predefined body locations: the left and right sides of the chest, waist, pocket, and ankle. Only frames in which the person is visible are retained for training and evaluation. Figure~\ref{fig:motiv_scene} shows an example RGB frame and the corresponding range–azimuth radar image, where brighter pixels indicate stronger reflections, typically from moving limbs or hidden metal objects.

\parlabel{Models.}  
We evaluate three state-of-the-art reconstruction-based anomaly detection models--GLAD~\cite{yao2024glad}, SiFall~\cite{ji2023sifall}, and CGAN~\cite{toma2020deep}---representing three dominant generative architectures: diffusion, variational autoencoder (VAE), and conditional GAN, respectively. These models operate by learning to reconstruct normal inputs and computing anomaly scores based on reconstruction error, a principle that has been shown to perform well across a variety of sensing modalities including vision, audio, and RF signals. Although some of these models were originally developed for other domains such as vision or WiFi sensing, we adapt them to our setting by modifying their input pipelines to accept mmWave radar spectra. Specifically, we convert raw radar data into range–azimuth spectra, resize and normalize them as required by each model, and retain their original architectures and anomaly scoring methods. Since mmWave-based reconstruction models are still an emerging area and existing models tailored for mmWave anomaly detection are limited, we select these top-performing methods to establish a strong baseline for our empirical study.

\begin{figure}[!t]
\centering
\includegraphics[width=0.96\linewidth]{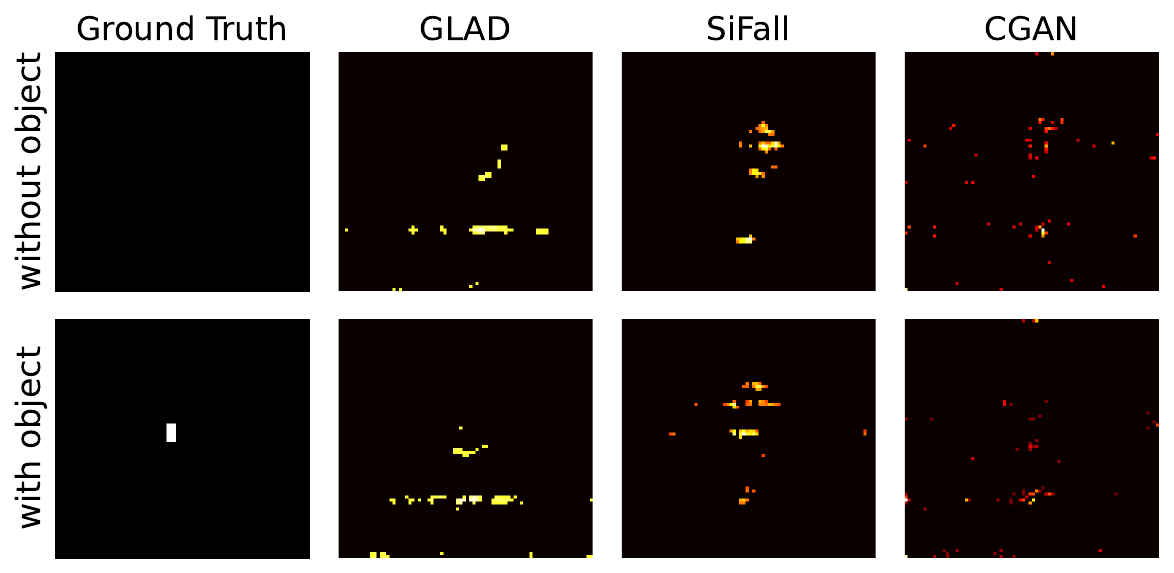}
\vspace{-3mm}
\caption{Anomaly maps generated by three baseline models using only mmWave radar data. Axes: x—Azimuth, y—Range. Brighter regions indicate higher anomaly scores.}
\label{fig:mm_novc}
\vspace{-3mm}
\end{figure}

\begin{figure}[!t]
\centering
\includegraphics[width=0.96\linewidth]{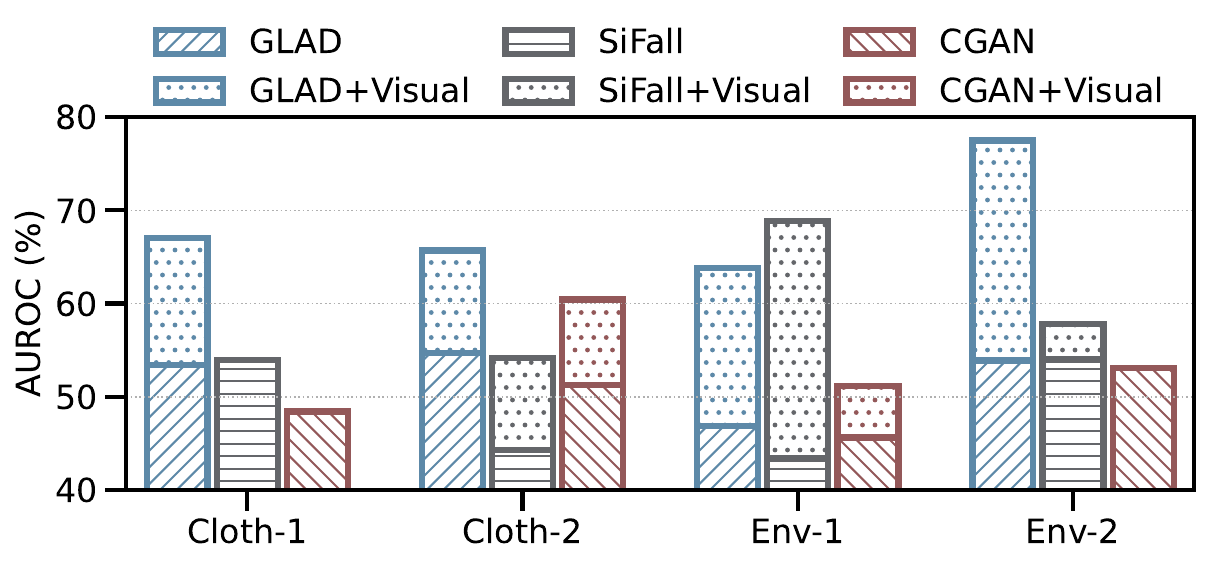}
\vspace{-3mm}
\caption{Accuracy of baseline anomaly detection models augmented with visual context.}
\label{fig:cond_results}
\vspace{-3mm}
\end{figure}

\begin{figure*}[!t]
\centering
\begin{subfigure}[c]{0.45\textwidth}
\centering
\includegraphics[width=\textwidth]{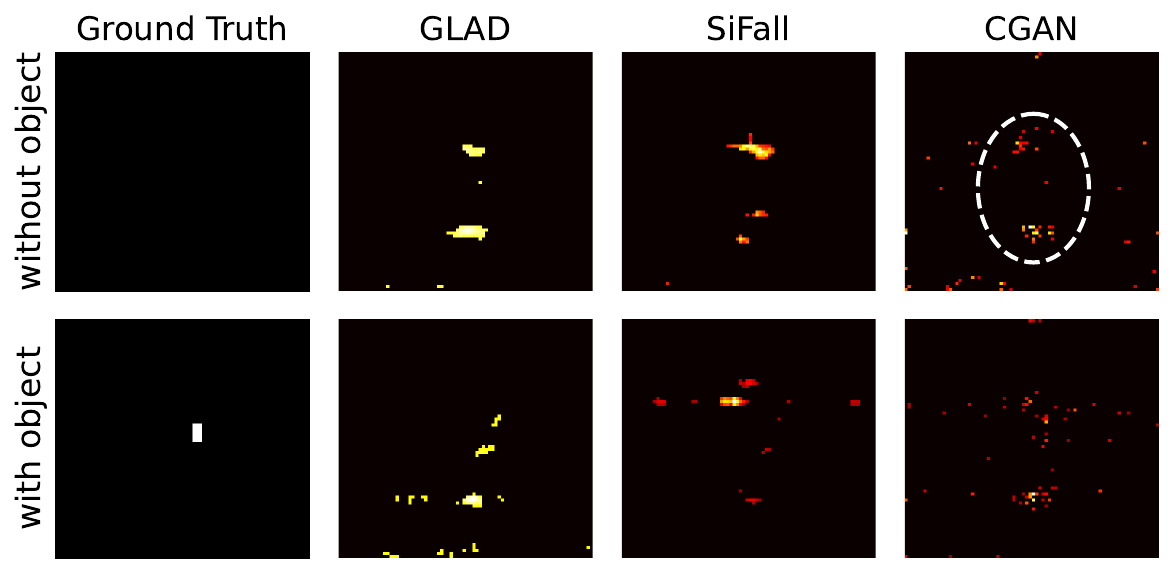}
\caption{Clothing-2 (no visual context)}
\label{fig:novc_snow_anomaly}
\end{subfigure}
\hfil
\begin{subfigure}[c]{0.45\textwidth}
\centering
\includegraphics[width=\textwidth]{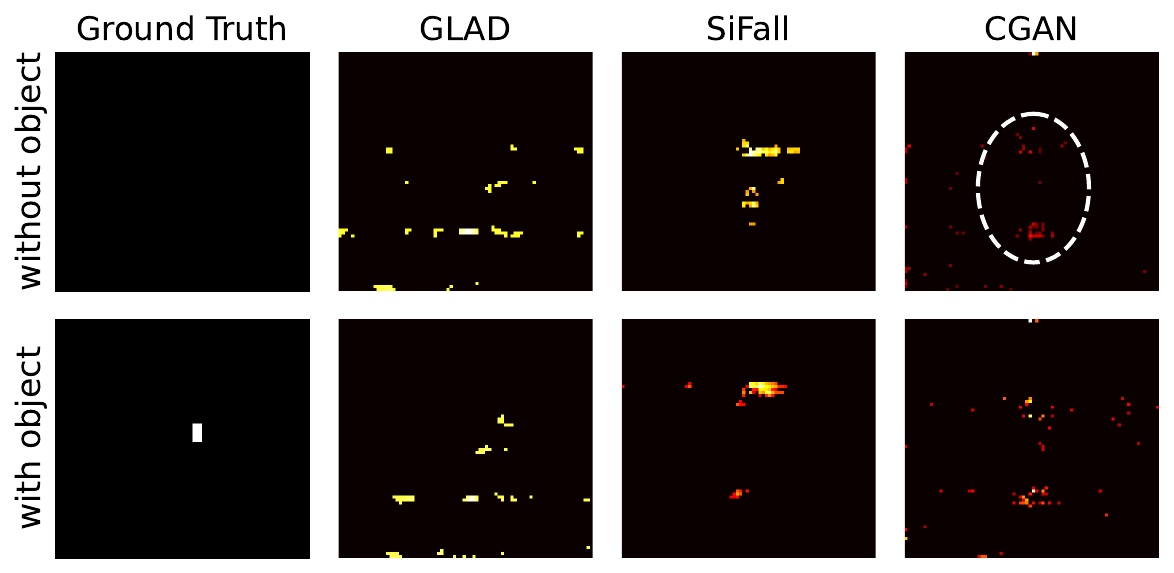}
\caption{Clothing-2 (with visual context)}
\label{fig:snow_anomaly}
\end{subfigure}
\hfil
\begin{subfigure}[c]{0.45\textwidth}
\centering
\includegraphics[width=\textwidth]{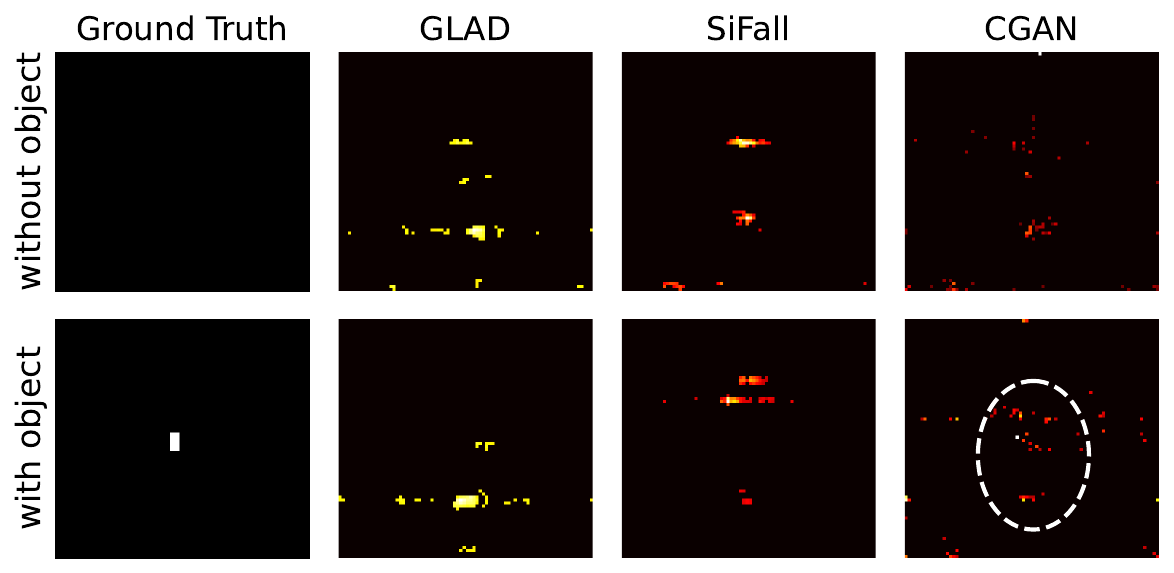}
\caption{Environment-1 (no visual context)}
\label{fig:novc_ladder_anomaly}
\end{subfigure}
\hfil
\begin{subfigure}[c]{0.45\textwidth}
\centering
\includegraphics[width=\textwidth]{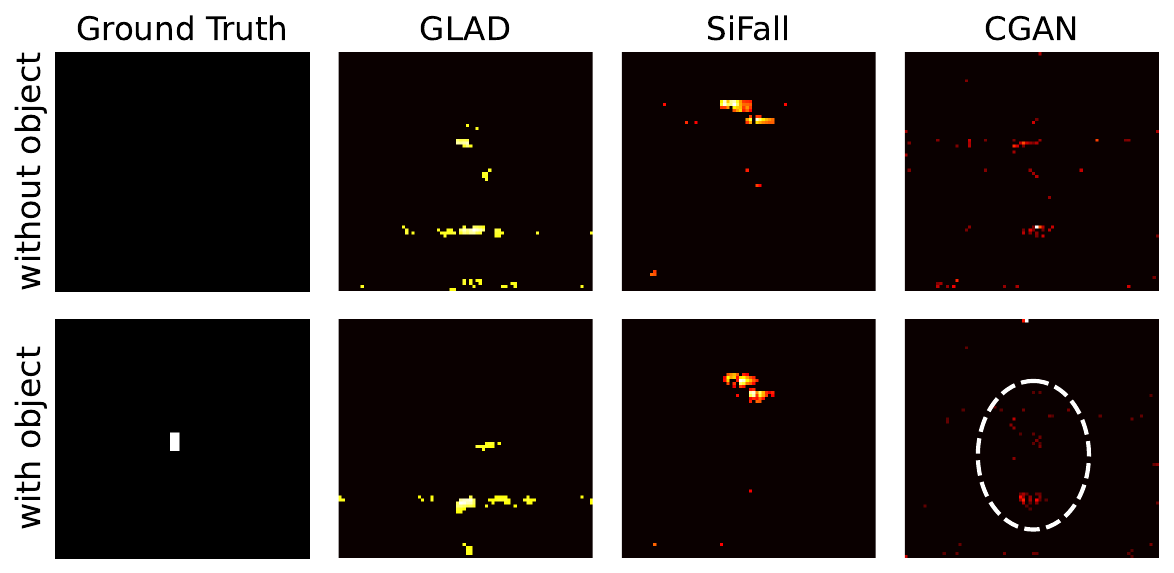}
\caption{Environment-1 (with visual context)}
\label{fig:ladder_anomaly}
\end{subfigure}
\caption{Anomaly maps generated by baseline models using mmWave radar and visual context. Axes: x—Azimuth, y—Range. Brighter regions indicate higher anomaly scores. White dashed circles highlight regions where false positives are reduced after incorporating visual context.}
\label{fig:mm_vc}
\end{figure*}

\parlabel{Training and Testing Protocol.}  
We follow the standard one-class paradigm employed by reconstruction-based detectors. Each model—GLAD (diffusion), SiFall (VAE), and CGAN (conditional GAN)—is trained solely on weapon-free spectra, which account for 70\,\% of the dataset. The networks thus learn the distribution of normal range–azimuth spectra: GLAD iteratively denoises to a clean reconstruction, SiFall decodes the VAE latent mean, and CGAN’s generator produces a conditional reconstruction.  

For \textbf{testing}, the remaining 30\% of recordings, containing both normal passages and spectra with a concealed metallic object in any of eight body regions, are processed frame-by-frame. Each model produces a reconstructed spectrum and computes an anomaly score based on the difference between the input and its reconstruction. The decision threshold is selected using the calibration procedure described in the corresponding original paper; we adopt the same percentile-based rule without modification. All other hyperparameters—loss functions, learning rate schedules, and preprocessing steps—follow the original papers.


\parlabel{Evaluation Metrics.}  
We evaluate performance using Area Under the Receiver Operating Characteristic Curve (AUROC), which measures the likelihood that an anomalous frame receives a higher anomaly score than a normal one. This is a standard metric in anomaly detection~\cite{liu2023simplenet, chen2024unified, yao2024glad, strater2024generalad}. We also report Average Precision (AP), which captures the area under the precision-recall curve and better reflects model performance on imbalanced datasets where anomalies are rare.

\subsection{mmWave-Only Anomaly Detection}  

This section compares anomaly detection results using only mmWave data, without any visual cues from RGB images.

\parlabel{Quantitative Results.}  
Figure~\ref{fig:soa_results} summarizes the anomaly detection performance of GLAD, SiFall, and CGAN using only mmWave inputs. GLAD performs best, with AUROC slightly above 50\% and AP near 83\%, followed by CGAN. SiFall shows the weakest performance across both metrics. These results suggest that diffusion-based reconstruction (GLAD) is more effective at modeling normal mmWave spectra and identifying deviations, whereas the VAE-based SiFall struggles to capture the underlying distribution. On average, AUROC across models is only 49\%, indicating poor reliability in ranking anomalous frames above normal ones. This suggests that models often assign higher anomaly scores to normal samples, leading to misclassification. In contrast, the average AP is 82\%, reflecting better precision-recall performance for the anomaly class. However, since AP does not penalize false positives on normal frames, the AUROC–AP gap reveals a key limitation: the models may detect some anomalies but also misclassify normal frames, showing limited generalization in the mmWave-only setting.

\parlabel{Anomaly Map Visualization.}  
Figure~\ref{fig:mm_novc} shows qualitative anomaly maps from GLAD, SiFall, and CGAN on mmWave spectra. Brighter regions indicate higher anomaly scores. The first column shows the ground truth, with the white pixel in the bottom row marking the concealed object. In the top row (no object), all models produce false positives—highlighting benign regions as anomalous—consistent with the low AUROC scores. In the bottom row (with object), the models assign elevated scores but fail to localize the actual anomaly. GLAD produces more structured activations, CGAN shows scattered highlights, and SiFall yields vertically diffuse responses. These visualizations reinforce earlier findings: while the models detect deviations, they struggle with precision—misclassifying normal frames and failing to isolate true anomalies—highlighting the challenge of mmWave-only anomaly detection without visual context.

\parlabel{Remarks.}  
The poor performance of existing models on mmWave radar data stems from three key limitations. First, most generative models assume Gaussian noise in the reconstruction space, which poorly reflects the structured, non-Gaussian distortions caused by multipath, occlusion, and material interactions in mmWave signals. Second, these models lack contextual awareness—they do not incorporate external cues such as clothing, posture, or environmental layout, all of which strongly influence radar reflections. As a result, they frequently misclassify benign variations as anomalies. Third, these models are unimodal and rely solely on radar data, without access to the visual scene. Without observing physical objects and their spatial context, they struggle to generate accurate, context-conditioned mmWave spectra.

\begin{figure*}[!t]
\centering
\includegraphics[width=\linewidth]{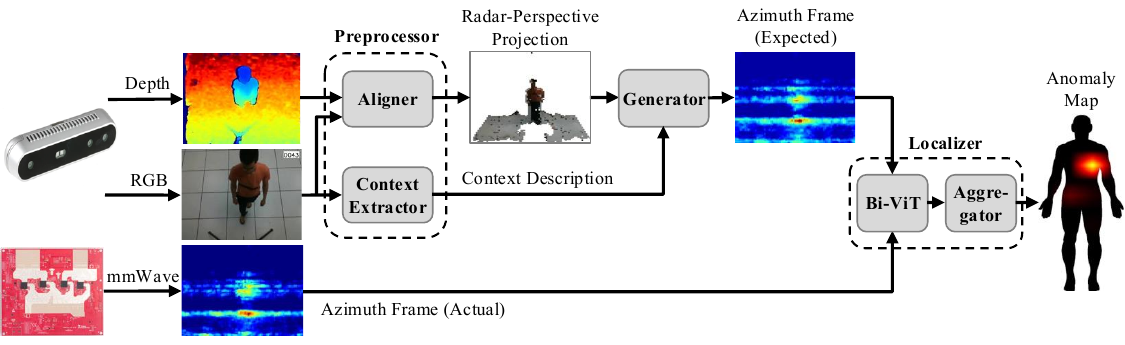}
\vspace{-3mm}
\caption{\sys Framework: The system takes synchronized RGB, depth, and Range-Azimuth spectrum inputs and produces an anomaly map that localizes anomalies in the mmWave domain. While the example shown illustrates concealed object localization on the human body, the framework is designed to detect and localize a broad range of non-visual anomalies.}
\label{fig:pix2mm_block}
\vspace{-3mm}
\end{figure*}

\subsection{Enhancing Detection with Visual Cues}  



 Although cameras cannot see concealed objects or anomalies observed by mmWave radar, they capture aspects of the scene—such as clothing type, material properties, and room geometry—that influence how mmWave signals propagate. By incorporating these visual cues into the mmWave-based anomaly detection pipeline, we can provide additional context that helps the model better reconstruct normal patterns and more reliably detect anomalies.


\parlabel{Use of Visual Cues.} To incorporate visual context into mmWave anomaly detection in our controlled empirical study, we assume oracle context and manually annotate visual attributes—such as clothing type, room layout, furniture, and material properties—by inspecting the ground-truth RGB images and metadata. This manual labeling is used only to isolate the effect of context on signal generation by eliminating errors from an imperfect context classifier. In the full \sys pipeline evaluated later in the paper, these attributes are obtained automatically via a learned context-classification module rather than manual or hard-coded labels. Rather than feeding visual features directly into the models, we group mmWave recordings by visual conditions and fine-tune pretrained models (GLAD, SiFall, CGAN) on weapon-free samples within each group. This context-specific fine-tuning provides a controlled and interpretable way to evaluate how visual cues influence the radar signal distribution.


\parlabel{Quantitative Results and Anomaly Maps.}  
Figure~\ref{fig:cond_results} shows that incorporating visual context consistently improves anomaly detection performance across all settings. For instance, GLAD's AUROC increases from 55\% to 66\% under Clothing-2 and from 47\% to 64\% under Environment-1, while SiFall shows even larger gains, reaching 69\% in the latter case. These improvements highlight the value of context-specific fine-tuning, which helps models better capture variations in radar signatures caused by clothing and environmental factors. The anomaly maps in Figure~\ref{fig:mm_vc} further illustrate these effects. Without visual cues, models often produce false positives in normal frames (e.g., Clothing-2) or misinterpret background clutter (e.g., Environment-1). With visual context, the reconstructions become more stable, and anomaly highlights are more focused around true object regions. GLAD tends to produce more focused and structured maps, especially with visual context, while SiFall often shows vertically diffuse activations and CGAN exhibits scattered responses. These patterns help explain the variation in detection performance across models and contexts.



\parlabel{Summary of Findings.} Both \textbf{model choice} and \textbf{visual context} play key roles in reconstruction error-based mmWave anomaly detection. Stronger generative models improve reconstruction quality, while visual context captures environmental and physical cues that radar alone misses—together enabling more robust, accurate, and context-aware detection.


%% file: tex/04_method.tex
\section{The \sys Framework}
\label{sec:overview}

\parlabel{Overview.} Building on the earlier findings that combining robust models with visual context significantly enhances anomaly detection, we propose \sys, a multimodal, cross-domain generative framework designed to detect and localize anomalies in non-visual scenarios. Examples include concealed weapon detection, intruders behind walls, and rare events such as falls occurring beyond visual obstacles. Unlike traditional vision-based methods, which cannot inherently sense these hidden or obstructed events, mmWave radar uniquely penetrates obstacles and senses subtle, invisible movements, making it ideally suited for detecting anomalies in non-visual domains.

To bridge these modalities, \sys leverages visual context from a co-located camera that captures key environmental details—such as object layout, clothing materials, human positions, and body configurations—that influence mmWave signal characteristics. Using this information, our cross-modal generative AI system predicts anomaly-free mmWave signals for normal activities, such as walking without concealed items, routine behavior behind walls, or stable posture without accidental falls. By comparing these synthesized signals with real radar observations, \sys identifies and localizes deviations that indicate anomalies, even in visually inaccessible scenarios.

\parlabel{Framework Illustration.} Figure~\ref{fig:pix2mm_block} illustrates the \sys framework using concealed weapon localization as an example scenario, where the goal is to identify the region of the human body associated with anomalous radar reflections. The system takes synchronized RGB, depth, and mmWave radar inputs. The RGB and depth streams are fused and reprojected into a \textit{Radar-Perspective Projection}, aligning the visual data with the radar’s field of view. This projection, along with a context description generated from the RGB input, is fed into the \textit{Generator}, which synthesizes an expected anomaly-free azimuth frame. The \textit{Localizer} then compares this with the actual radar frame to produce an anomaly map highlighting the location of the concealed object.

\parlabel{Functional Modules.} The overall architecture of \sys is composed of three core modules: \textit{Preprocessor}, \textit{Generator}, and \textit{Localizer}. Each module performs a distinct transformation that enables cross-modal anomaly detection:

\begin{itemize}[leftmargin=10pt]
\item \textbf{Preprocessor:} Combines two submodules—(i) the \textit{Aligner}, which fuses RGB and depth inputs into a \textit{Radar-Perspective Projection} aligned with the radar’s Range–Azimuth view, and (ii) the \textit{Context Extractor}, which derives high-level semantic cues (e.g., clothing material, environment) and composes them into a natural-language prompt that conditions radar spectrum generation.

\item \textbf{Generator:} Receives the radar-perspective projection and the context prompt, and synthesizes an expected anomaly-free mm-Wave spectrum using a cross-modal latent diffusion model. This serves as a baseline representation of normalcy for comparison.

\item \textbf{Localizer:} Compares the synthesized and observed mmWave spectra using a dual-branch Vision Transformer and outputs an anomaly map that localizes deviations, whether on the human body or elsewhere in the scene.

\end{itemize}

\subsection{Preprocessor Module}
\label{sec:preprocessor}

\parlabel{Rationale.}
The \textit{Preprocessor} module integrates two key functions, spatial alignment and semantic context extraction, to ensure that visual information is meaningfully comparable with mmWave radar signals. First, the \textit{Aligner} fuses RGB and depth frames and reprojects them into the radar’s coordinate system, producing a \textit{Radar-Perspective Projection} that is spatially consistent with the radar’s Range–Azimuth map. This enables the system to correlate visual structures (e.g., body outlines, obstacles) with corresponding radar reflections. 

Second, the \textit{Context Extractor} analyzes the RGB input to identify high-level semantic cues, such as clothing materials and environmental features, and encodes them into a structured natural-language prompt. These cues influence mmWave propagation and are essential for conditioning the subsequent radar spectrum generation. Together, these two steps provide both geometric alignment and semantic grounding, allowing the system to generate reliable anomaly-free radar predictions and accurately localize anomalies.

\begin{figure}[!t]
\centering
\includegraphics[width=\linewidth]{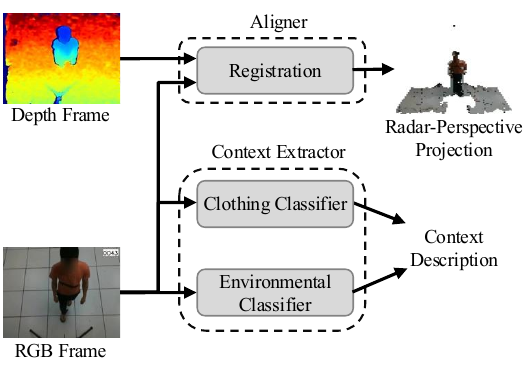}
\vspace{-6mm}
\caption{Architecture of the \textit{Preprocessor} module: The Aligner submodule reprojects RGB-depth inputs into the radar’s perspective and the Context Extractor submodule derives semantic cues to generate a descriptive prompt.}
\label{fig:context_block}
\vspace{-3mm}
\end{figure}

\parlabel{Processing Steps.}
At each timestamp \( t \), the Preprocessor takes synchronized RGB \( C_t \), depth \( D_t \), and radar \( R_t \) inputs and performs two complementary operations: (i) geometric alignment of the RGB–depth inputs to the radar’s coordinate system, and (ii) semantic extraction of visual context relevant to mmWave propagation. The unified processing pipeline (Figure~\ref{fig:context_block}) consists of the following steps:

\begin{itemize}[leftmargin=10pt]

\item \parlabel{Step 1: RGB–Depth Geometry Alignment.}
The depth map is first converted into a 3D point cloud using camera intrinsics. These 3D points are then projected back onto the RGB plane for consistent correspondence. Finally, the point cloud is reprojected into the radar’s coordinate space and discretized into range bins to form a 2D \textit{Radar-Perspective Projection} aligned with the Range–Azimuth map.

\item \parlabel{Step 2: Visual Context Classification.}
From the RGB frame, two ResNet classifiers independently predict:  
(i) the \textit{clothing material} worn by the subject, and  
(ii) the \textit{environment type} (e.g., corridor, stairwell).  
Both are trained using standard cross-entropy loss: $\mathcal{L}_{CE} = \log\!\left(\sum_{j=0}^{C-1} e^{z_j}\right) - z_c$, where \( \mathbf{z} \in \mathbb{R}^C \) are the predicted logits and \( c \) is the true class label.

\item \parlabel{Step 3: Natural-Language Prompt Construction.}
The predicted clothing and environmental labels are combined into a structured context prompt that describes the expected radar behavior in the current scene. The prompt template is: \texttt{\small ``A person is wearing [clothing-phrase], which [clothing-effect]. The person is walking toward the radar in an environment where [environment-effect]. Generate the expected radar spectrum assuming no anomalies."} This prompt conditions the downstream diffusion-based radar generator.

\end{itemize}

\parlabel{Implementation.}
We implement the \textit{Preprocessor} using two components. The \textit{Aligner}, built with a Cython-accelerated utility on the Intel RealSense SDK~\cite{intel2025align}, aligns the depth map to the RGB frame, reconstructs a 3D point cloud, and projects it into the radar’s coordinate space to produce the \textit{Radar-Perspective Projection}. The \textit{Context Extractor} uses two lightweight ResNet-18 classifiers~\cite{he2016deep} for clothing and environment recognition, trained for 150 epochs in PyTorch~2.0 on a single NVIDIA V100 GPU (16\,GB). Together, these modules provide geometrically aligned and semantically enriched visual context for downstream spectrum generation.

\subsection{Generator Module}
\label{sec:generator}

\parlabel{Rationale.}
The \textit{Generator} module is responsible for synthesizing an expected, anomaly-free mmWave spectrum that reflects how radar signals should appear under normal conditions, given the visual and semantic context of the scene. This expected spectrum serves as a baseline for detecting deviations in the actual radar signal and plays a central role in identifying anomalies.

The Generator is a cross-modal generative AI module that takes two inputs: the radar-perspective projection from the Aligner and the descriptive prompt from the Context Extractor. These inputs encode both spatial features and high-level semantic cues—such as clothing type and environmental layout—that significantly influence radar reflections but are not directly observable in the raw radar data. By conditioning on both modalities, the Generator can produce a contextually accurate reference signal that reflects the expected radar behavior for the current scene.

We employ a vision-language conditional generative model for this purpose, which enables robust generation without requiring explicit anomaly labels. This design allows the system to learn generalized notions of normalcy across diverse environments and human conditions.

For example, given a prompt describing a person wearing a thick coat and walking indoors, the Generator produces a radar spectrum consistent with the reduced reflectivity and signal spread caused by such clothing and setting. If the actual radar signal deviates significantly from this generated baseline—say, due to a concealed object—it can then be flagged by the downstream Localizer module as a potential anomaly.

\begin{figure}[!t]
\centering
\includegraphics[width=\linewidth]{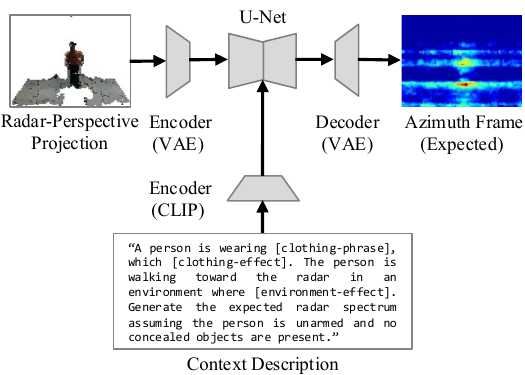}
\vspace{-6mm}
\caption{Architecture of the Generator module: The module performs cross-modal spectrum generation using a latent diffusion model conditioned on both the visual context and the textual prompt.}
\label{fig:diffusion_block}
\vspace{-3mm}
\end{figure}

\parlabel{Processing Steps.}  
As shown in Figure~\ref{fig:diffusion_block}, our generation pipeline is based on a latent diffusion model for image-to-image translation~\cite{parmar2024pix2pix} and proceeds through the following steps:

\begin{itemize}[leftmargin=10pt]

\item \parlabel{Step 1: Latent Encoding.}  
The radar-perspective visual context image is encoded into a compact latent representation using a convolutional encoder, reducing dimensionality while preserving structural information.

\item \parlabel{Step 2: Text Encoding.}  
The corresponding textual prompt, generated from scene semantics, is embedded using a pretrained CLIP encoder~\cite{radford2021learning}, providing a semantic conditioning vector for guided generation.

\item \parlabel{Step 3: Denoising with U-Net.}  
The latent visual embedding and semantic text embedding are fused and passed through a U-Net denoising network, which iteratively refines the latent representation to synthesize the expected mmWave spectrum.

\item \parlabel{Step 4: Decoding.}  
The denoised latent output is decoded back into pixel space, yielding the final generated radar spectrum aligned with the given visual and semantic context.

\item \parlabel{Step 5: Training Objective.}  
The generator is trained end-to-end using a composite loss function \( \mathcal{L}_{\text{gen}} \), which integrates multiple complementary objectives to guide fidelity, realism, distributional alignment, and semantic consistency.

\end{itemize}

\parlabel{Loss Function.}  
The generator is trained to minimize a composite loss: $\mathcal{L}_{\text{gen}} = \mathcal{L}_{\text{sim}} + \mathcal{L}_{\text{div}} + \mathcal{L}_{\text{adv}} + \mathcal{L}_{\text{text}}$, where the terms respectively encourage perceptual similarity, distributional alignment, visual realism, and semantic consistency.

\begin{itemize}[leftmargin=10pt]

\item \parlabel{Similarity Loss (\( \mathcal{L}_{\text{sim}} \)).}  
Encourages the generated spectrum to resemble the ground truth both perceptually and structurally. It combines multiple image similarity metrics: $\mathcal{L}_{\text{sim}} = \lambda_{\text{MSE}} \mathcal{L}_{\text{MSE}} + \lambda_{\text{LPIPS}} \mathcal{L}_{\text{LPIPS}} + \lambda_{\text{MSSIM}} \mathcal{L}_{\text{MSSIM}} + \lambda_{\text{DISTS}} \mathcal{L}_{\text{DISTS}}$

\item \parlabel{Divergence Loss (\( \mathcal{L}_{\text{div}} \)).}  
Promotes distributional alignment between generated and real spectra via KL divergence: $\mathcal{L}_{\text{div}} = \lambda_{\text{KLD}} \mathcal{L}_{\text{KLD}}$

\item \parlabel{Adversarial Loss (\( \mathcal{L}_{\text{adv}} \)).}  
Encourages realism in the generated output through a GAN-based discriminator: $\mathcal{L}_{\text{adv}} = \lambda_{\text{GAN}} \mathcal{L}_{\text{GAN}}$

\item \parlabel{Textual Loss (\( \mathcal{L}_{\text{text}} \)).}  
Ensures the generated spectrum is semantically aligned with the input prompt using CLIP-based vision-language similarity: $\mathcal{L}_{\text{text}} = \lambda_{\text{CLIP}} \mathcal{L}_{\text{CLIP}}$

\end{itemize}

\parlabel{Implementation.}  
We implement the Generator module by extending the open-source conditional stable diffusion framework from~\cite{parmar2025img2img}. Our implementation incorporates a customized generation loss \( \mathcal{L}_{\text{gen}} \), combining similarity, divergence, adversarial, and textual alignment objectives. Perceptual and structural losses are implemented using the PyTorch Image Quality (PIQ) library~\cite{kastryulin2022piq}. The model is trained for over 100 epochs, with checkpoints saved every 2 epochs. All training is conducted on a single NVIDIA A100 GPU with 40 GB of memory.

\subsection{Localizer Module}
\label{sec:localizer}

\parlabel{Rationale.}
The \textit{Localizer} module is responsible for detecting and localizing anomalies by comparing the real mmWave signal with the expected, anomaly-free signal generated in the previous stage. The goal is to identify spatial regions where radar reflections deviate from what is predicted under normal visual and semantic context—such deviations may indicate concealed objects, abnormal behavior, or other non-visual anomalies.

To achieve this, the module takes as input the real and generated Range-Azimuth spectra and processes them using a dual-branch Vision Transformer (ViT) architecture. Each branch independently encodes one of the two spectra, and their features are then jointly analyzed to detect localized differences. This comparison enables the system to isolate anomalous patterns in the radar signal while remaining invariant to normal variations caused by scene or subject diversity. The output of this module is a fine-grained anomaly map that spatially localizes the source of irregularities, supporting both detection and interpretability.

For example, if the generated spectrum reflects a walking person but the real radar shows extra torso reflections, the Localizer flags this discrepancy as an anomaly—potentially indicating a concealed object.

\parlabel{Processing Steps.}  
The anomaly localization pipeline proceeds through the following stages:

\begin{itemize}[leftmargin=10pt]

\item \parlabel{Step 1: Patch Encoding.}  
The real and generated mmWave spectra are divided into fixed-size patches and independently passed through two Vision Transformer (ViT) encoders~\cite{dosovitskiy2021image}, each producing a sequence of patch embeddings.

\item \parlabel{Step 2: Feature Extraction.}  
From each encoder, we extract the CLS token, which serves as a global summary of the corresponding input spectrum.

\item \parlabel{Step 3: Feature Fusion.}  
The two CLS tokens are concatenated and passed through a linear projection layer to obtain a joint latent representation that captures discrepancies between the real and generated signals.

\item \parlabel{Step 4: Classification.}  
The fused representation is classified into one of \( C \) predefined anomaly classes, each corresponding to a specific spatial region of the human body. A softmax activation yields class-wise anomaly probabilities.

\item \parlabel{Step 5: Multi-Frame Aggregation.}  
To ensure temporal stability, majority voting is applied over sequential predictions from adjacent frames, producing a final class label for the current frame.

\item \parlabel{Step 6: Anomaly Map Generation.}  
The predicted class label is visualized as a spatial anomaly map, highlighting suspicious regions on the body and aiding interpretability.

\end{itemize}

\parlabel{Loss Function.}  
The Localizer is trained using a composite loss function that combines representation-level and classification objectives:

\begin{itemize}[leftmargin=10pt]

\item \parlabel{MSE Loss (\( \mathcal{L}_{\text{MSE}} \)).}  
This loss measures the discrepancy between the real and generated CLS token embeddings using mean squared error:  $\mathcal{L}_{\text{MSE}} = \frac{1}{N} \sum_{i=1}^N (\hat{y}_i - y_i)^2$, where \( \hat{y}_i \) and \( y_i \) are the CLS token vectors from the generated and real branches, and \( N \) is the embedding dimension.

\item \parlabel{Cross-Entropy Loss (\( \mathcal{L}_{\text{CE}} \)).}  
Standard cross-entropy loss is applied to the predicted anomaly class label. This loss formulation is consistent with the classification loss described in Section~\ref{sec:preprocessor}.

\item \parlabel{Localization Loss (\( \mathcal{L}_{\text{loc}} \)).}  
The overall training objective combines the two losses:  $\mathcal{L}_{\text{loc}} = \lambda_{\text{MSE}} \mathcal{L}_{\text{MSE}} + \lambda_{\text{CE}} \mathcal{L}_{\text{CE}}$, where \( \lambda_{\text{MSE}} \) and \( \lambda_{\text{CE}} \) are tunable hyperparameters that balance the importance of representation alignment and classification accuracy.

\end{itemize}

\parlabel{Implementation.}  
We implement the Localizer using a dual-branch Vision Transformer (ViT) architecture based on the ViT-B/16 variant~\cite{dosovitskiy2021image}, adapted from the open-source implementation by~\cite{jeon2025vit}. The model is initialized with pretrained weights from ImageNet-21K~\cite{ridnik2021imagenet21k}. We modify the architecture to accept real and generated spectra as parallel input streams and add a classification head optimized using our composite localization loss \( \mathcal{L}_{\text{loc}} \). The training setup—including batching, optimizer settings, and validation routines—follows the same protocol used for the Preprocessor.

\begin{figure}[!t]
\centering
\includegraphics[width=\linewidth]{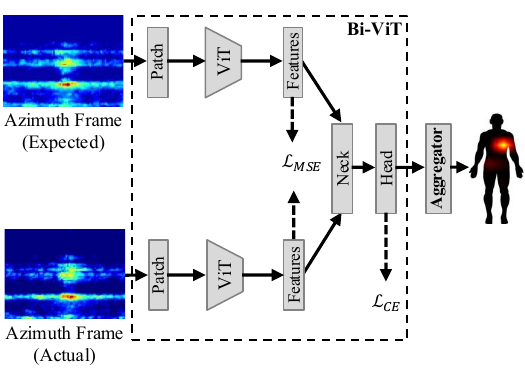}
\vspace{-6mm}
\caption{Architecture of the Localizer module: The module detects and localizes anomalies by comparing real and generated spectra through a dual-branch Vision Transformer (ViT) encoder followed by an Aggregator.}
\label{fig:multi_vit_block}
\vspace{-3mm}
\end{figure}

%% file: tex/05_evalutaion.tex
\section{Microbenchmarks}

\label{sec:experimental-eval}

We evaluate the core components of \sys—Preprocessor, Generator, and Localizer—through targeted experiments measuring context extraction, spectrum generation quality, and localization accuracy. These results validate each module’s effectiveness and set the stage for the next section, which evaluates \sys in complete application scenarios.

\subsection{Experimental Setup}
We evaluate the three modules of \sys using five large and diverse public datasets: TextileNet~\cite{zhong2023textilenet}, HuPR~\cite{lee2023hupr}, UWCR~\cite{gao2021rampcnn}, mmCounter~\cite{toha2025mmcounter}, and Carry Object~\cite{gao2024mmwcarry}. TextileNet provides over 240k avatar images with material-level fiber labels across 32 classes (e.g., cotton, wool, polyester), supporting the modeling of clothing-dependent mmWave propagation. HuPR contains 141k paired RGB-mmWave frames of a single person performing static actions, waving, and walking. UWCR offers 19k RGB–mmWave pairs from complex on-road scenes featuring cars, trucks, motorbikes, cycles, and pedestrians. mmCounter includes 7k RGB–mmWave samples of multi-person indoor scenarios with varying group configurations. Carry Object provides 1.8k RGB–mmWave pairs for concealed/open carry detection of knives, keys, laptops, and mobile phones.

To evaluate robustness under inaccurate multimodal alignment, we augment the Carry Object dataset using four perturbation types, yielding a total of 9k RGB–mmWave pairs: (i) \textit{Gaussian Blur} to simulate motion or lens defocus, (ii) \textit{Lighting and Contrast} adjustments to mimic auto-exposure variation, (iii) \textit{Additive Gaussian Noise} to emulate camera sensor noise, and (iv) \textit{Random Shift (Misalignment)} to simulate radar–camera calibration errors.

\bnote{These datasets are used to evaluate different modules of the system as follows: (i) the TextileNet dataset is used to evaluate the Preprocessor module, (ii) the HuPR, UWCR, mmCounter, and Carry Object datasets are used to evaluate the Generator module, and (iii) the Carry Object dataset is used to evaluate the Localizer module.}


\subsection{Evaluation of Preprocessor Module}
\label{sec:preprocess_eval}

The goal of this experiment is to evaluate the accuracy–latency trade-off of the neural networks used in the Preprocessor module and determine which model is best suited for real-time deployment in \sys. We compare three image classifiers—VGG-11, ResNet-50, and ViT-B/16—using the TextileNet dataset. We measure two quantities: (i) top-5 classification accuracy (reported using the F1 score), and (ii) inference latency per batch of 256 images. The only varying factor across experiments is the network architecture.

\begin{figure}[!t]
\centering
\begin{subfigure}[c]{0.23\textwidth}
\centering
\includegraphics[width=\textwidth]{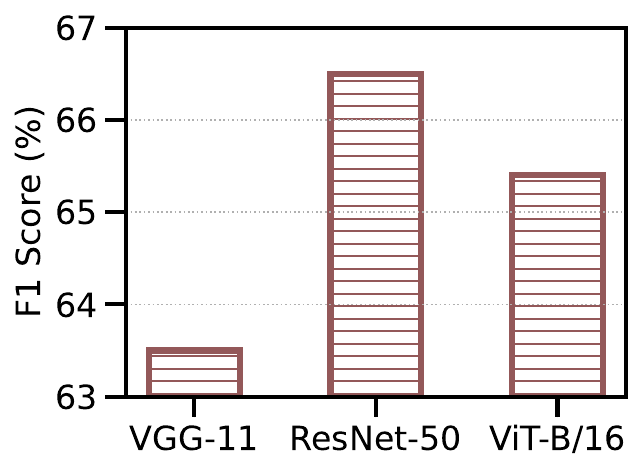}
\caption{F1 Score ($\uparrow$ is better)}
\end{subfigure}
\hfil
\begin{subfigure}[c]{0.23\textwidth}
\centering
\includegraphics[width=\textwidth]{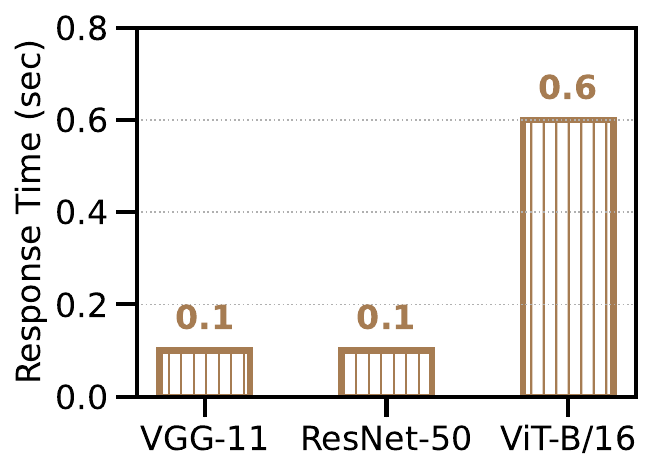}
\caption{Response time ($\downarrow$ is better)}
\end{subfigure}
\vspace{-3mm}
\caption{Evaluation of Preprocessor module}
\label{fig:preprocess_eval}
\vspace{-3mm}
\end{figure}

As shown in Figure~\ref{fig:preprocess_eval}, the CNN-based ResNet-50 achieves the highest accuracy (F1 $\sim$67\%) while also maintaining the lowest latency ($\sim$0.1\,\text{s}). VGG-11 exhibits similar latency but lower accuracy ($\sim$63\%), consistent with its known limitations related to vanishing gradients. In contrast, the self-attention-based ViT-B/16 achieves higher accuracy than VGG-11 but incurs substantially higher computation cost. Overall, ResNet-50 provides the best balance of accuracy and efficiency, making it the preferred choice for real-time preprocessing in \sys.

\subsection{Evaluation of Generator Module}
\label{sec:gen_eval}

The goal of this experiment is to evaluate how well \sys generates high-quality, context-aware mmWave spectra compared to two generative baselines. We compare \sys against CVAE and CGAN across four public datasets—HuPR, UWCR, mmCounter, and Carry Object. CVAE is adapted from the SiFall model using a PyTorch VAE~\cite{anand2025vae}, while CGAN is based on Pix2Pix~\cite{phillipi2025pix2pix}. \bnote{CVAE and CGAN are not multi-modal and therefore do not receive any text prompt, whereas \sys includes a multi-modal generator that conditions on the context description via a text prompt.} The experiment varies only the generative model architecture and measures two metrics: distributional realism using FID \cite{heusel2017gans} (lower is better) and perceptual similarity using DISTS \cite{ding2022unifying} (higher is better). 

\begin{figure}[!t]
\centering
\begin{subfigure}[c]{0.46\textwidth}
\centering
\includegraphics[width=\textwidth]{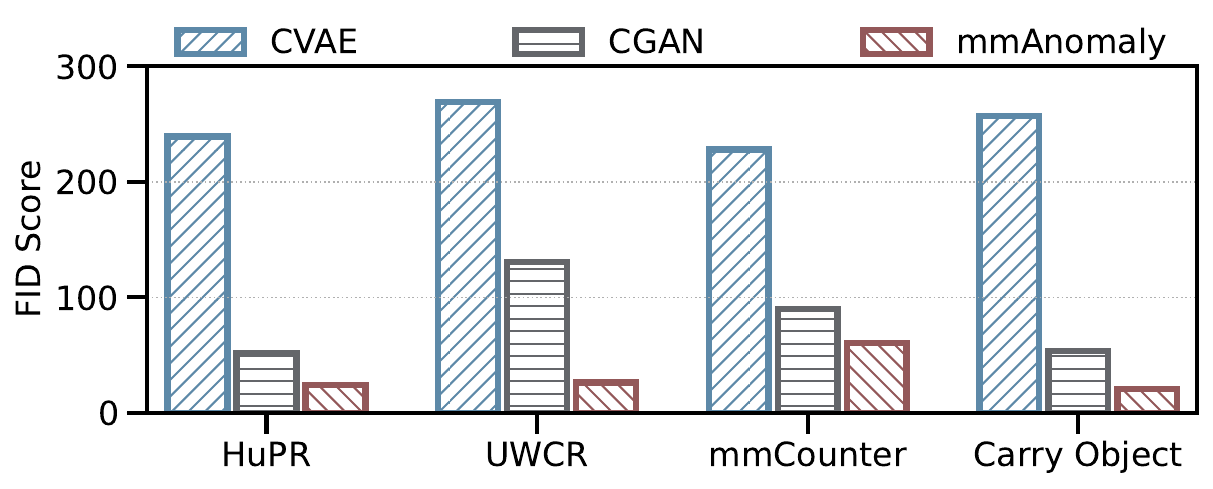}
\caption{Fr\'echet Inception Distance ($\downarrow$ better)}
\end{subfigure}
\hfil
\begin{subfigure}[c]{0.46\textwidth}
\centering
\includegraphics[width=\textwidth]{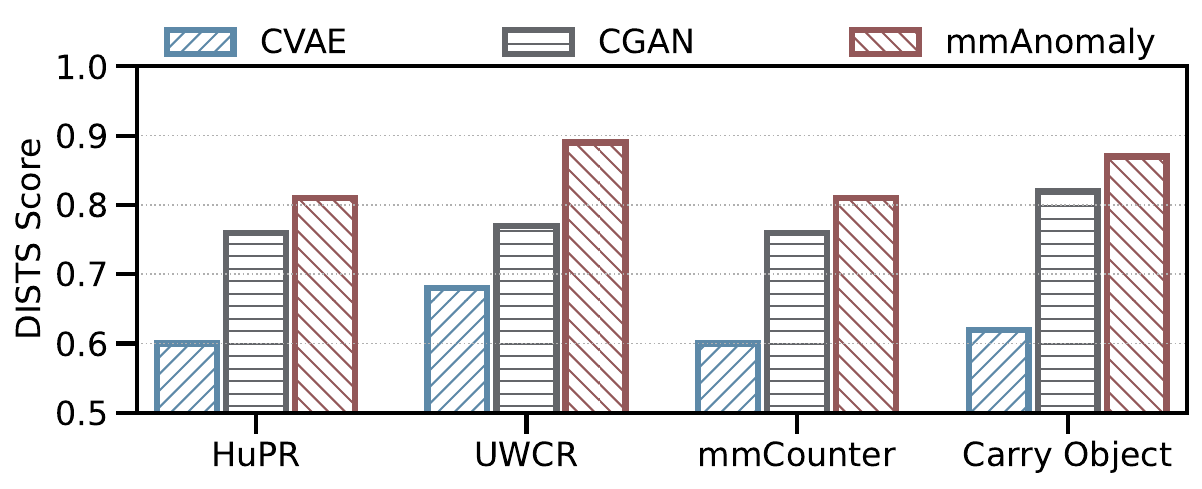}
\caption{DISTS Score ($\uparrow$ better)}
\end{subfigure}
\vspace{-3mm}
\caption{Evaluation of Generator module}
\label{fig:gen_eval}
\vspace{-3mm}
\end{figure}

As shown in Figure~\ref{fig:gen_eval}, \sys consistently achieves the best performance across all datasets, exhibiting both higher DISTS and lower FID than the baselines. CVAE performs the worst on both metrics, while CGAN yields moderate performance, aligning with expectations from prior generative modeling literature. Overall, these results demonstrate that the context-guided diffusion generator in \sys produces radar spectra with superior visual fidelity and distributional accuracy, confirming the effectiveness of our cross-modal generation design.

\begin{figure}[!t]
\centering
\begin{subfigure}[c]{0.23\textwidth}
\centering
\includegraphics[width=\textwidth]{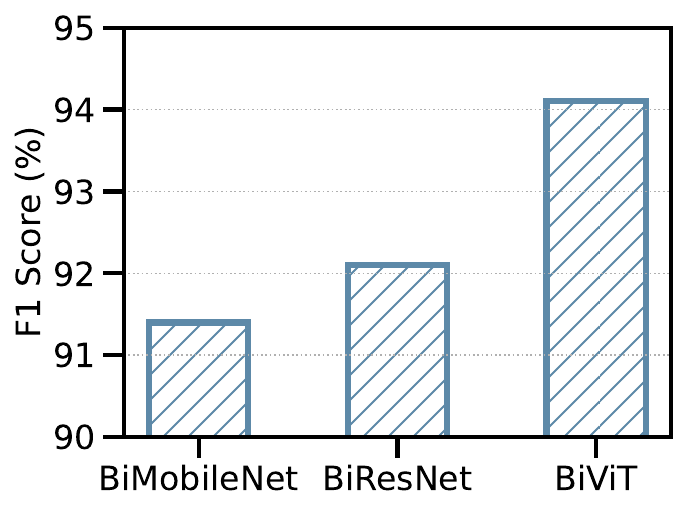}
\caption{Multi-class F1 Score ($\uparrow$ better)}
\end{subfigure}
\hfil
\begin{subfigure}[c]{0.23\textwidth}
\centering
\includegraphics[width=\textwidth]{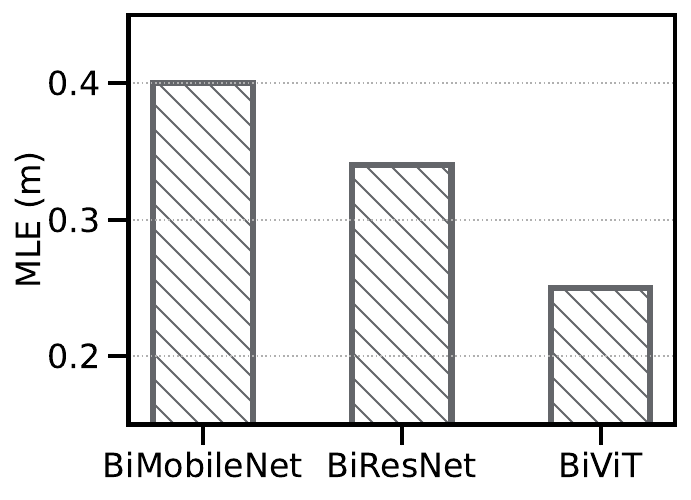}
\caption{Mean localization error ($\downarrow$ better)}
\end{subfigure}
\vspace{-3mm}
\caption{Evaluation of Localizer module}
\label{fig:loc_eval}
\vspace{-3mm}
\end{figure}

\subsection{Evaluation of Localizer Module}
\label{sec:loc_eval}

The goal of this experiment is to evaluate how well \sys localizes anomalies by comparing real and generated spectra, and to assess whether its self-attention–based design outperforms CNN-based bimodal architectures. We compare \sys against two baselines—BiMobileNet and BiResNet—using the Carry Object dataset. To create controlled anomalies, we synthesize anomalous radar spectra by injecting ghost objects into random regions of real spectra. The experiment varies only the bimodal architecture and measures two metrics: the F1 score and the mean localization error.

As shown in Figure~\ref{fig:loc_eval}, \sys (BiViT) consistently achieves the highest F1 score and the lowest localization error. While BiMobileNet and BiResNet are lightweight CNN-based models, they fail to achieve comparable localization performance and exhibit higher localization error across all trials. Overall, these findings demonstrate that the self-attention–based bimodal Localizer in \sys more accurately pinpoints mmWave anomalies than CNN-based alternatives, confirming the effectiveness of our Localizer design.

%% file: tex/06_applications.tex
\section{Case Studies}

We present three case studies demonstrating the practical utility of \sys: through-cloth weapon detection, through-wall intrusion detection, and fall detection behind walls.

\subsection{Through-Cloth Weapon Localization}
\label{sec:weapon-det}

\parlabel{Experimental Setup.}
We train \sys pipeline over the same self-collected concealed weapon dataset discussed in Section~\ref{sec:empirical}. For comparison, we use the same anomaly detection baselines—CVAE and CGAN (both multimodal)—as in Section~\ref{sec:experimental-eval}. We also include a non-generative radar-only baseline, MMW-Carry~\cite{gao2024mmwcarry}, adapted for direct classification. Localization performance is evaluated using multi-class F1 score (higher is better) across eight body regions, and Mean Localization Error (MLE, lower is better) for spatial precision in millimeters. \bnote{To calculate the MLE, we use fixed physical distances between the eight canonical body regions derived from the average human body in our dataset (i.e., averaged across all participants).}

\begin{figure}[!t]
\centering
\begin{subfigure}[c]{0.23\textwidth}
\centering
\includegraphics[width=\textwidth]{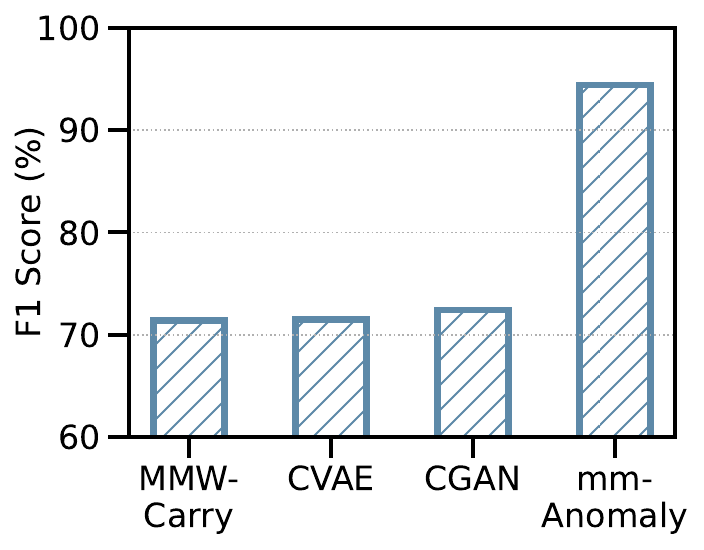}
\caption{Multi-class F1 Score ($\uparrow$ better)}
\end{subfigure}
\hfil
\begin{subfigure}[c]{0.23\textwidth}
\centering
\includegraphics[width=\textwidth]{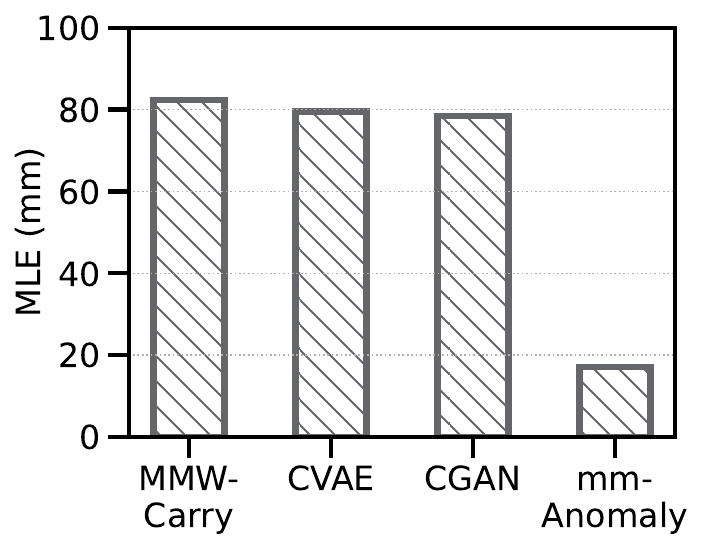}
\caption{Mean localization error ($\downarrow$ better)}
\end{subfigure}
\vspace{-3mm}
\caption{Weapon detection and localization accuracy of \sys against the baseline models}
\label{fig:weapon_eval}
\vspace{-3mm}
\end{figure}

\begin{figure}[!t]
\centering
\begin{subfigure}[c]{0.23\textwidth}
\centering
\includegraphics[width=\textwidth]{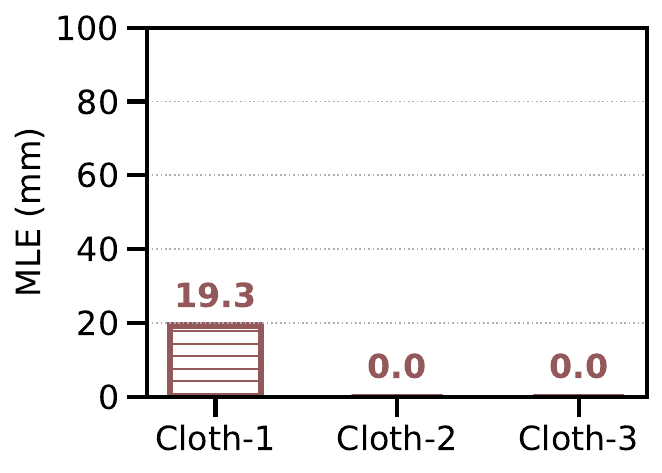}
\caption{Impact of clothing materials}
\end{subfigure}
\hfil
\begin{subfigure}[c]{0.23\textwidth}
\centering
\includegraphics[width=\textwidth]{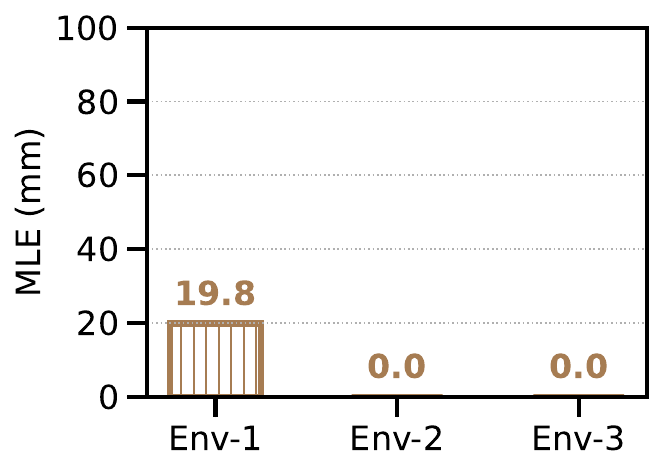}
\caption{Impact of environmental setups}
\end{subfigure}
\vspace{-3mm}
\caption{Weapon localization accuracy of \sys for seen clothing materials and environmental setups}
\label{fig:cloth_env_eval}
\vspace{-3mm}
\end{figure}

\parlabel{Weapon Detection and Localization Accuracy.} 
We evaluate \sys's ability to detect and localize weapons concealed under clothing. As shown in Figure~\ref{fig:weapon_eval}, \sys outperforms all three baselines with an F1 score of $\sim$94\% and a mean localization error (MLE) of $\sim$16\,mm, compared to 72–73\% F1 and 78–82\,mm MLE for the baselines. On average, this reflects a 23\% gain in accuracy and a 63\,mm improvement in localization precision. These gains are attributed to \sys's context-aware spectrum generation and dual-branch transformer architecture, which improve both anomaly detection reliability and spatial resolution.

\parlabel{Impact of Clothing Materials.} 
We evaluate \sys's localization robustness across three clothing conditions: casual dress (Cloth-1), fleece jacket (Cloth-2), and snow jacket (Cloth-3). As shown in Figure~\ref{fig:cloth_env_eval}, \sys maintains low mean localization error in all cases, demonstrating strong generalization to varying garment types and material thicknesses. A slightly higher error in Cloth-1 is due to the greater variability in casual attire, though all errors remain below 2\,cm.

\parlabel{Impact of Environmental Setups.}  
We evaluate \sys’s localization accuracy across three environments: a typical lab (Env-1), with a ladder (Env-2), and with both ladder and whiteboard (Env-3). As shown in Figure~\ref{fig:cloth_env_eval}, \sys maintains consistently low localization error across all conditions, confirming its robustness to environmental clutter. Env-1 shows slightly higher error due to more frequent human movement during data collection, yet all localization errors remain below 2\,cm.

\begin{figure}[!t]
\centering
\begin{subfigure}[c]{0.23\textwidth}
\centering
\includegraphics[width=\textwidth]{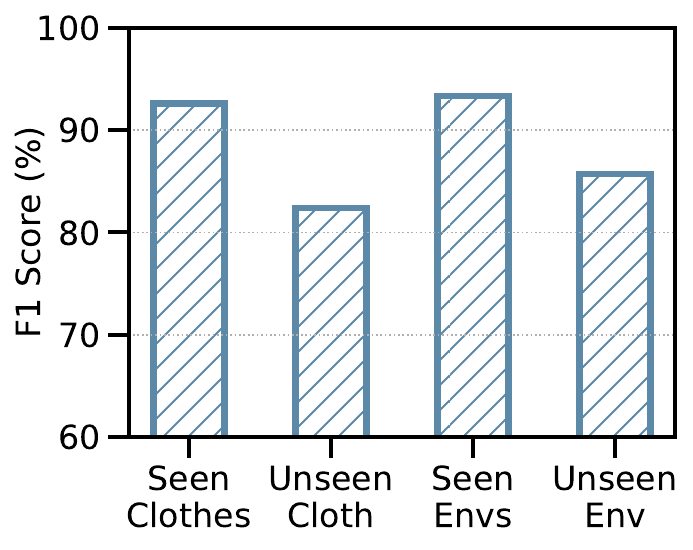}
\caption{Multi-class F1 Score ($\uparrow$ better)}
\end{subfigure}
\hfil
\begin{subfigure}[c]{0.23\textwidth}
\centering
\includegraphics[width=\textwidth]{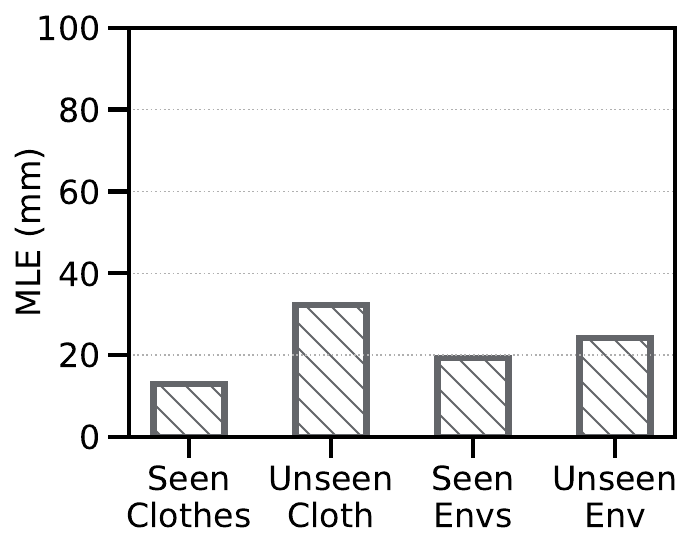}
\caption{Mean localization error ($\downarrow$ better)}
\end{subfigure}
\vspace{-3mm}
\caption{Weapon localization accuracy of \sys for unseen clothing materials and environmental setups}
\label{fig:unseen_eval}
\vspace{-3mm}
\end{figure}

\parlabel{Impact of Unseen Visual Cues.}
We evaluate \sys under both unseen clothing and unseen environment scenarios. In the unseen clothing evaluation, \sys is trained on three clothing types and tested on the remaining one. Similarly, in the unseen environment evaluation, \sys is trained on three environments and tested on the held-out environment. As shown in Figure~\ref{fig:unseen_eval}, \sys maintains a high F1 score and achieves a low mean localization error under unseen conditions, confirming strong generalization performance.

\begin{figure}[!t]
\centering
\begin{subfigure}[c]{0.23\textwidth}
\centering
\includegraphics[width=\textwidth]{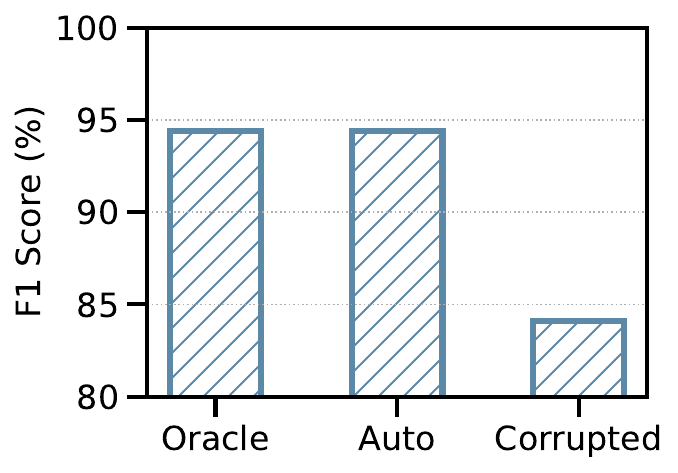}
\caption{Multi-class F1 Score ($\uparrow$ better)}
\end{subfigure}
\hfil
\begin{subfigure}[c]{0.23\textwidth}
\centering
\includegraphics[width=\textwidth]{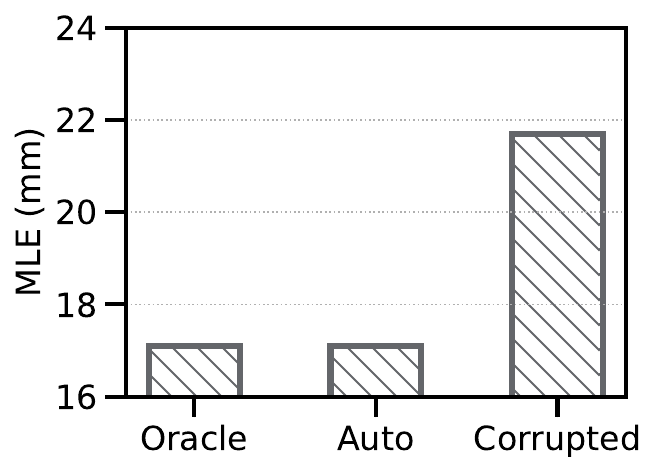}
\caption{Mean localization error ($\downarrow$ better)}
\end{subfigure}
\vspace{-3mm}
\caption{Sensitivity analysis of context quality on the through-cloth weapon dataset}
\label{fig:oracle_eval}
\vspace{-3mm}
\end{figure}

\parlabel{Sensitivity to Context Quality.}
\bnote{We analyze the sensitivity of \sys to context quality using the through-cloth weapon dataset. To this end, we modify the preprocessing stage to produce three types of context inputs: (i) oracle context (ground-truth labels), (ii) automatically predicted context, and (iii) intentionally corrupted context. Each variant is passed through the same Generator--Localizer pipeline to predict weapon locations. As shown in Figure~\ref{fig:oracle_eval}, automatically predicted context (\sys) achieves nearly identical F1 score and mean localization error compared to the oracle setting, indicating that the Preprocessor module provides sufficiently accurate visual cues for reliable downstream detection.}

\subsection{Through-Wall Intruder Localization}
\label{sec:intrusion-det}
\begin{figure}[!t]
\centering
\includegraphics[width=\linewidth]{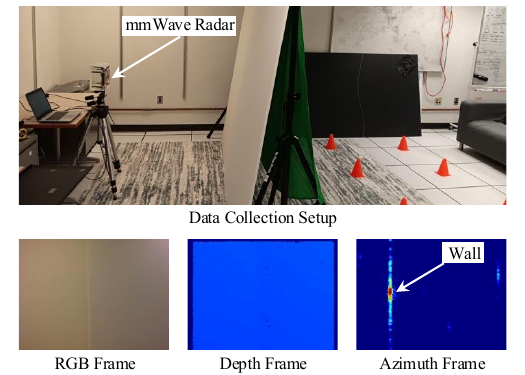}
\vspace{-6mm}
\caption{Through-wall data collection setup. The RGBD camera cannot see through the wall, but the radar can.}
\label{fig:wall_setup}
\vspace{-4mm}
\end{figure}

\parlabel{Dataset Preparation.} We collect a multi-modal dataset for detecting and localizing intruders behind walls using a mmWave radar and an RGBD camera mounted on a 4-foot tripod with a fixed tilt of \ang{0}. The setup faces seven wall configurations, including four materials—curtain, gator board, particle board, and styrofoam—and two reflective objects: a ladder and a poster board. For each scene, we record a blank case (no person) as the normal condition, followed by intruder scenarios where five participants stand or sit at one of six predefined positions behind the wall. The camera captures only the opaque wall, while the radar penetrates it to sense hidden human presence. The dataset contains 272 recordings, each with 60 frames at 2 fps. Figure~\ref{fig:wall_setup} shows the data collection setup. The RGB frame shows the wall, the depth frame shows the distance map, and Azimuth frame shows the top view of the room.

\parlabel{Experimental Setup.} We execute the end-to-end \sys pipeline for this task. For context extraction, we train a ResNet-18 classifier to identify the seven wall types, using the same configuration as the clothing and environment classifiers. To guide spectrum generation, we construct prompts using the template: \texttt{\small{``The radar is directed at a [wall-type], which [wall-effect]. There is no human or object present behind the wall. Generate the expected radar spectrum for this blank scene."}} We use several consecutive frames in the aggregation step to decide on the intruder's presence with location. 

For comparison, we use three localization baselines: RTWLBR \cite{wang2023realtime}, CVAE, and CGAN. RTWLBR is a mmWave-based through-Wall human localization approach, while CVAE and CGAN are deep generative approaches. Evaluation uses multi-class F1 score (higher is better) for classification accuracy across six location classes, and Mean Localization Error (MLE, lower is better) for spatial precision in meters.

\begin{figure}[!t]
\centering
\begin{subfigure}[c]{0.23\textwidth}
\centering
\includegraphics[width=\textwidth]{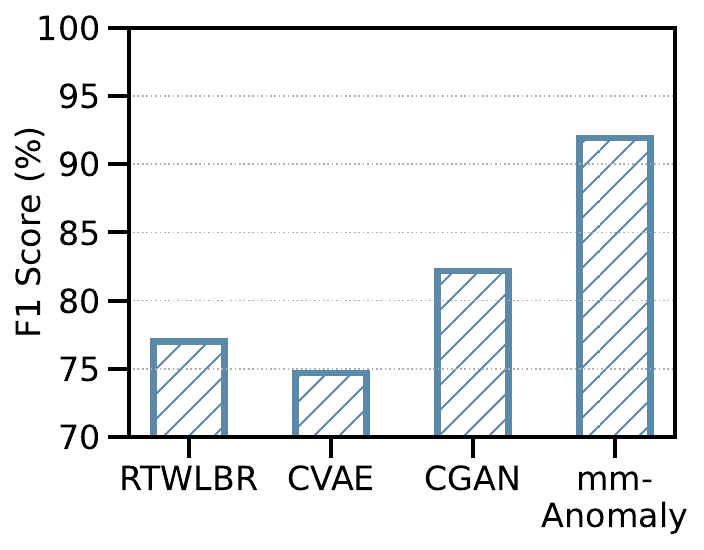}
\caption{Multi-class F1 Score ($\uparrow$ better)}
\end{subfigure}
\hfil
\begin{subfigure}[c]{0.23\textwidth}
\centering
\includegraphics[width=\textwidth]{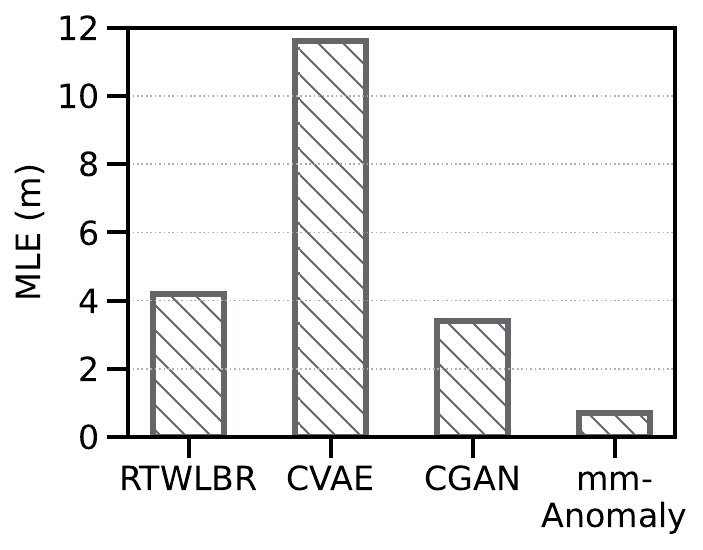}
\caption{Mean localization error ($\downarrow$ better)}
\end{subfigure}
\vspace{-3mm}
\caption{Intrusion detection and localization accuracy of \sys against the baseline models}
\label{fig:int_eval}
\vspace{-3mm}
\end{figure}

\parlabel{Intruder Detection and Localization Accuracy.} We evaluate \sys's ability to detect and localize human intruders through various wall materials. As shown in Figure~\ref{fig:int_eval}, \sys outperforms all baselines with an F1 score of $\sim$92\% (vs. 75–82\%) and the lowest MLE ($\sim$0.7\,m, vs. up to 11.6\,m for CVAE). On average, \sys improves F1 by $\sim$14\% and reduces localization error by $\sim$6\,m. These gains demonstrate \sys's strong generalization in occluded settings, enabled by its context-aware spectrum synthesis and cross-modal localization design.

\begin{figure}[!t]
\centering
\begin{subfigure}[c]{0.23\textwidth}
\centering
\includegraphics[width=\textwidth]{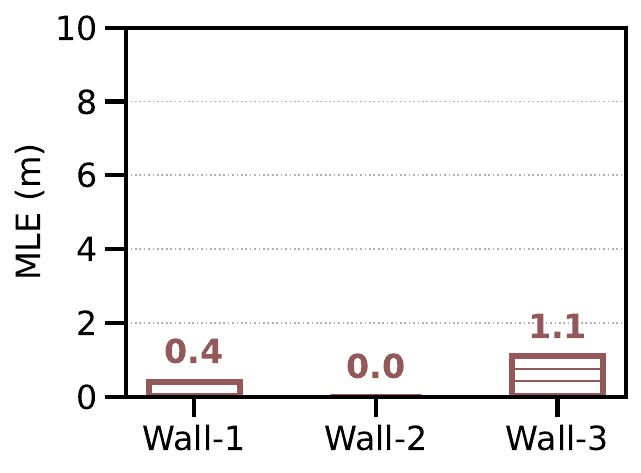}
\caption{Impact of wall materials}
\end{subfigure}
\hfil
\begin{subfigure}[c]{0.24\textwidth}
\centering
\includegraphics[width=\textwidth]{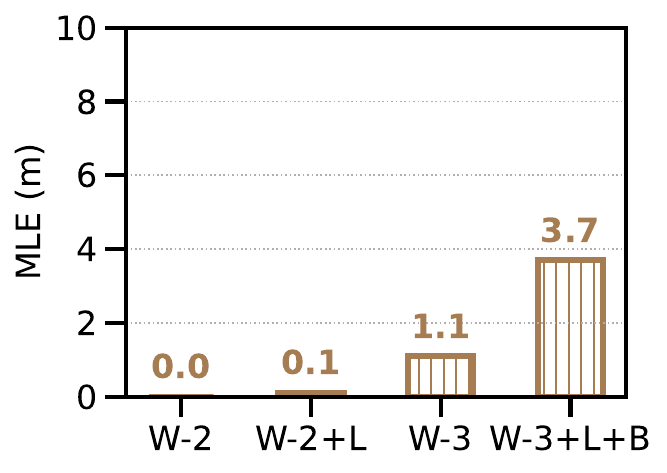}
\caption{Impact of reflectors}
\end{subfigure}
\vspace{-3mm}
\caption{Intrusion localization accuracy of \sys for different wall materials and reflectors}
\label{fig:wall_reflector_eval}
\vspace{-3mm}
\end{figure}

\parlabel{Impact of Wall Materials and Reflectors.}  
We evaluate \sys's localization robustness across wall materials—styrofoam (Wall-1), particle board (Wall-2), and gator board (Wall-3)—and four extended setups with added reflectors. As shown in Figure~\ref{fig:wall_reflector_eval}, \sys achieves near-zero MLE for most walls, showing strong generalization across structural barriers. Error increases in cluttered scenes with metal and board reflectors due to multipath effects, but \sys remains accurate under moderate clutter.

\subsection{Through-Wall Fall Localization}
\label{sec:fall_det}
\parlabel{Dataset.}
We use the self-collected multi-modal dataset from Section~\ref{sec:intrusion-det}, where \note{seven} volunteers simulate non-fall activities such as standing and sitting, and fall activities by transitioning from standing to sitting or lying down.

\parlabel{Experimental Setup.} We reuse the intruder localization pipeline for fall detection. As in intruder detection, \sys is trained to generate mmWave signals under normal conditions with no human present behind the wall using the visual cues. During inference, the Localizer identifies deviations from the generated spectrum, pinpointing the presence and location of a human as an anomaly. These localized anomalies, over time, form a sequence of spatial positions, which we use to infer movement patterns. The Localizer also predicts the pose of the human, such as sitting and standing. To detect falls, we compute the Shannon entropy of both pose and location time series in the aggregation step. A fall is detected when pose transitions from standing to sitting or lying, or when the human's location abruptly shifts or disappears—e.g., due to lying flat and becoming undetectable by radar. The Localization module uses these pose-aware locations to classify the event as normal or a fall. 

We use four baselines for comparison, such as Shen et al.~\cite{shen2024advanced}, MMFall~\cite{li2022realtime}, CVAE, and CGAN. Shen et al. proposed a conventional signal-processing-based fall detection method, and MMFall proposed an end-to-end deep-learning-based fall detection method. CVAE and CGAN are deep generative approaches, which we adopted for both fall detection and localization. Evaluation metrics include the multi-class F1 score (higher is better) for fall-detection accuracy and the Mean Localization Error (MLE, lower is better) for spatial precision.

\begin{figure}[!t]
\centering
\begin{subfigure}[c]{0.25\textwidth}
\centering
\includegraphics[width=\textwidth]{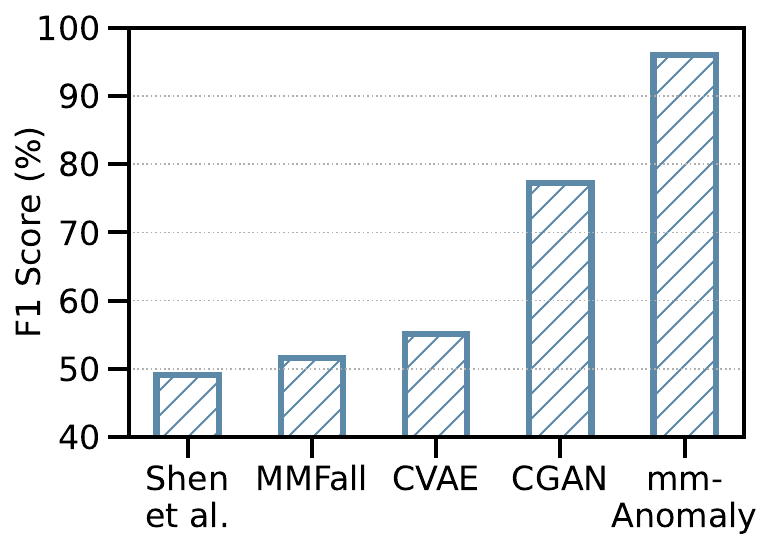}
\caption{Multi-class F1 Score ($\uparrow$ better)}
\end{subfigure}
\hfil
\begin{subfigure}[c]{0.22\textwidth}
\centering
\includegraphics[width=\textwidth]{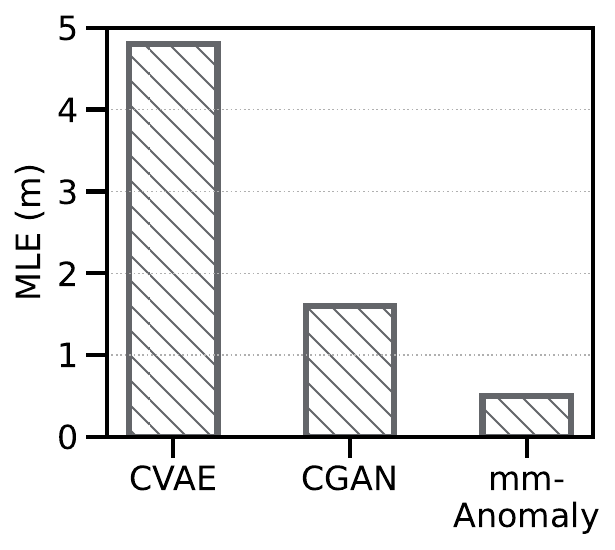}
\caption{Mean localization error ($\downarrow$ better)}
\end{subfigure}
\vspace{-3mm}
\caption{Fall detection and localization accuracy of \sys against the baseline models}
\label{fig:fall_eval}
\vspace{-3mm}
\end{figure}

\begin{figure}[!t]
\centering
\begin{subfigure}[c]{0.24\textwidth}
\centering
\includegraphics[width=\textwidth]{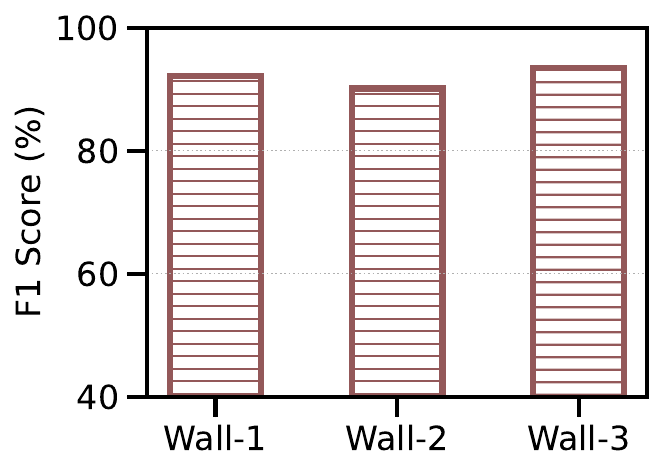}
\caption{Multi-class F1 score ($\uparrow$ better)}
\end{subfigure}
\hfil
\begin{subfigure}[c]{0.23\textwidth}
\centering
\includegraphics[width=\textwidth]{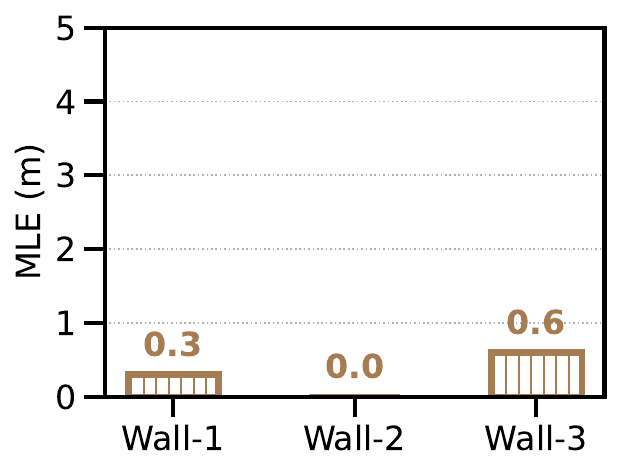}
\caption{Mean localization error ($\downarrow$ better)}
\end{subfigure}
\vspace{-3mm}
\caption{Fall detection and localization accuracy of \sys for different wall materials}
\label{fig:fall_wall_eval}
\vspace{-3mm}
\end{figure}

\parlabel{Fall Detection and Localization Accuracy.}  
We evaluate \sys’s effectiveness in detecting and localizing falls through walls, comparing it with CVAE and CGAN. As shown in Figure~\ref{fig:fall_eval}, \sys achieves the highest F1 score ($\sim$96\%) and lowest MLE ($\sim$0.5\,m), outperforming all baselines in both accuracy and spatial precision. On average, it improves F1 by $\sim$38\% and reduces localization error by $\sim$3\,m. These gains reflect \sys’s ability to capture motion transitions and radar signal loss through its pose-aware localization and context-guided generation.

\parlabel{Impact of Wall Materials.}  
We evaluate \sys’s robustness in fall detection across different wall materials. As shown in Figure~\ref{fig:fall_wall_eval}, \sys consistently achieves high F1 scores (93–97\%) and maintains MLE below 0.6\,m, with zero error for Wall-2. These results demonstrate reliable detection and localization performance regardless of wall composition.

\subsection{Application-Agnostic Evaluation}
\label{sec:misc_eval}

\begin{figure}[!t]
\centering
\begin{subfigure}[c]{0.23\textwidth}
\centering
\includegraphics[width=\textwidth]{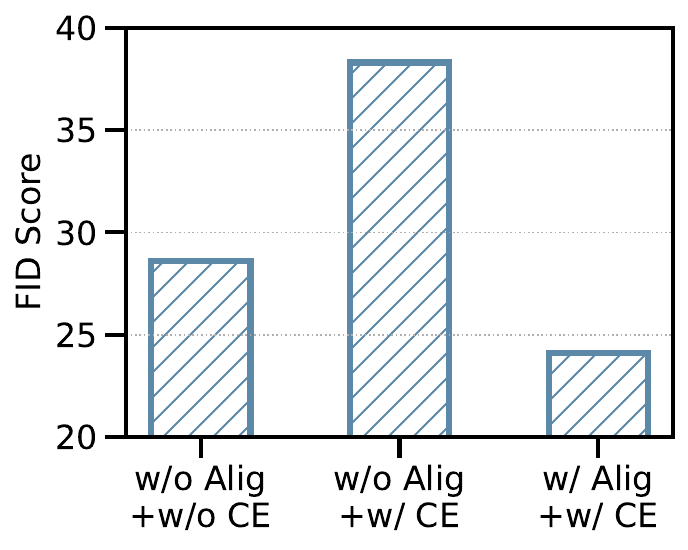}
\caption{Through-Cloth Dataset}
\label{fig:weapon_fid}
\end{subfigure}
\hfil
\begin{subfigure}[c]{0.23\textwidth}
\centering
\includegraphics[width=\textwidth]{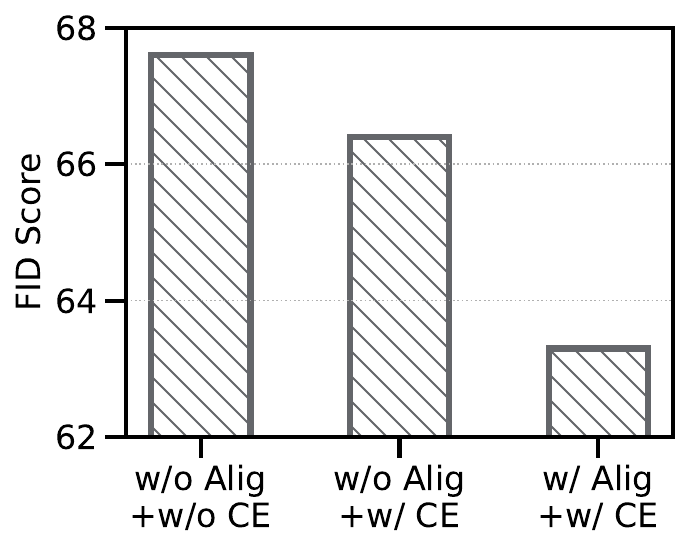}
\caption{Through-Wall Dataset}
\label{fig:wall_fid}
\end{subfigure}
\vspace{-3mm}
\caption{Ablating Aligner (Alig) and Context Extractor (CE) submodules using FID Score ($\downarrow$ better)}
\label{fig:ablation_fid}
\vspace{-3mm}
\end{figure}

\parlabel{Ablation Study.}
We perform an ablation study on the Aligner and Context Extractor using two self-collected datasets: through-cloth and through-wall. For the through-cloth dataset, removing the Aligner isolates the effect of human positional alignment, while removing the Context Extractor evaluates the contribution of clothing-related cues. Similarly, for the through-wall dataset, ablating the Aligner reflects the impact of wall-relative positioning, and ablating the Context Extractor isolates the role of wall material information. \bnote{To disable the Aligner, we feed the Generator with raw RGB inputs instead of radar-aligned projections; to disable the Context Extractor, we use a fixed prompt for all samples. As shown in Figure~\ref{fig:ablation_fid}, enabling both submodules consistently yields the lowest Fr\'echet Inception Distance across both datasets, indicating more realistic and context-consistent mmWave spectrum generation.}

\begin{figure}[!t]
\centering
\begin{subfigure}[c]{0.23\textwidth}
\centering
\includegraphics[width=\textwidth]{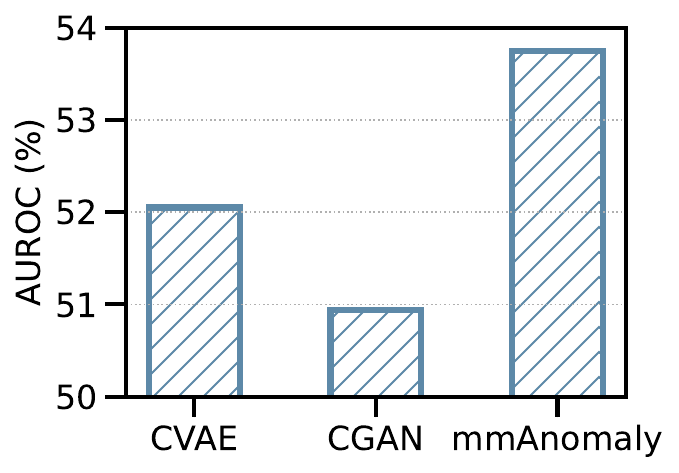}
\caption{Through-Cloth Dataset}
\end{subfigure}
\hfil
\begin{subfigure}[c]{0.23\textwidth}
\centering
\includegraphics[width=\textwidth]{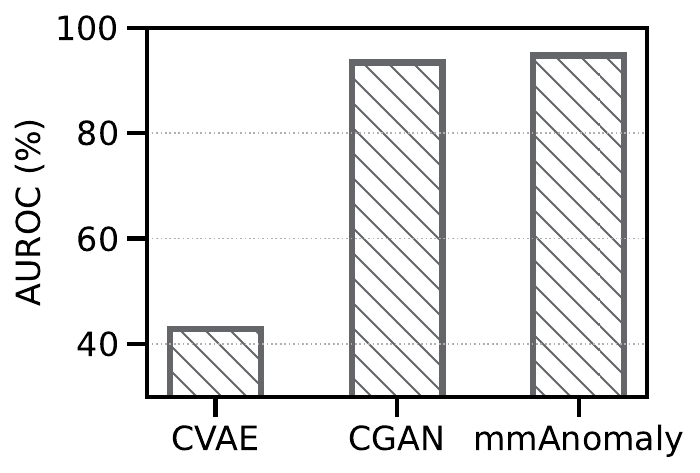}
\caption{Through-Wall Dataset}
\end{subfigure}
\vspace{-3mm}
\caption{Anomaly detection accuracy of \sys against the baseline models using AUROC Score ($\uparrow$ better)}
\label{fig:auroc_eval}
\vspace{-3mm}
\end{figure}

\parlabel{Anomaly Detection Accuracy.}
We evaluate the generic anomaly-detection capability of \sys by comparing it with CVAE and CGAN using AUROC on two self-collected datasets. AUROC quantifies how effectively a model separates anomalous samples from normal ones based on discrepancies between the generated and observed mmWave spectra. As shown in Figure~\ref{fig:auroc_eval}, \sys consistently outperforms both baselines, achieving modest gains on the through-cloth dataset and more pronounced improvements on the through-wall dataset. \bnote{These results indicate improved separability between normal and anomalous samples, enabling lower false-positive rates when selecting an operating threshold.} The smaller gain in the through-cloth setting is expected, as mmWave propagation is influenced by multiple physical factors even in nominal conditions, leading to benign signal variations that limit AUROC improvements.

\parlabel{Response Time Analysis.}
We evaluate the end-to-end inference time on an 80 GB NVIDIA H100 GPU using our self-collected through-cloth dataset. We break down the latency at the task level and show the results in Figure~\ref{fig:response_time}. Our findings show that \sys localizes concealed objects within one second, with the Aligner and Generator contributing the most to the overall latency. \bnote{Future optimization directions include replacing dense 3D reconstruction with Gaussian Splatting for faster RGB-D processing~\cite{niu2025anicrafter} and distilling the diffusion-based generator into a compact model~\cite{haoailab2026fastvideo} to reduce inference latency and GPU memory usage.}

\parlabel{Out-of-Distribution Contexts.}
To handle out-of-distribution contexts, we perform a multimodal foundation-model-based evaluation on the through-cloth dataset. We use the Gemini~2.5~Pro API to analyze RGB video and infer clothing materials and environmental reflectors relevant to mmWave propagation. On average, a single Gemini query takes $11.7 \pm 1.4$ seconds to process one RGB sequence. Although Gemini produces rich contextual descriptions, this latency makes it unsuitable for real-time use. \bnote{As a lightweight alternative for OOD handling, future work could incorporate open-set classification~\cite{carion2020detr} into the Context Extractor, allowing unseen clothing types or environments to be labeled as \emph{unknown} rather than causing false positives.}

%% file: tex/07_discussion.tex
\begin{figure}[!t]
\centering
\includegraphics[width=\linewidth]{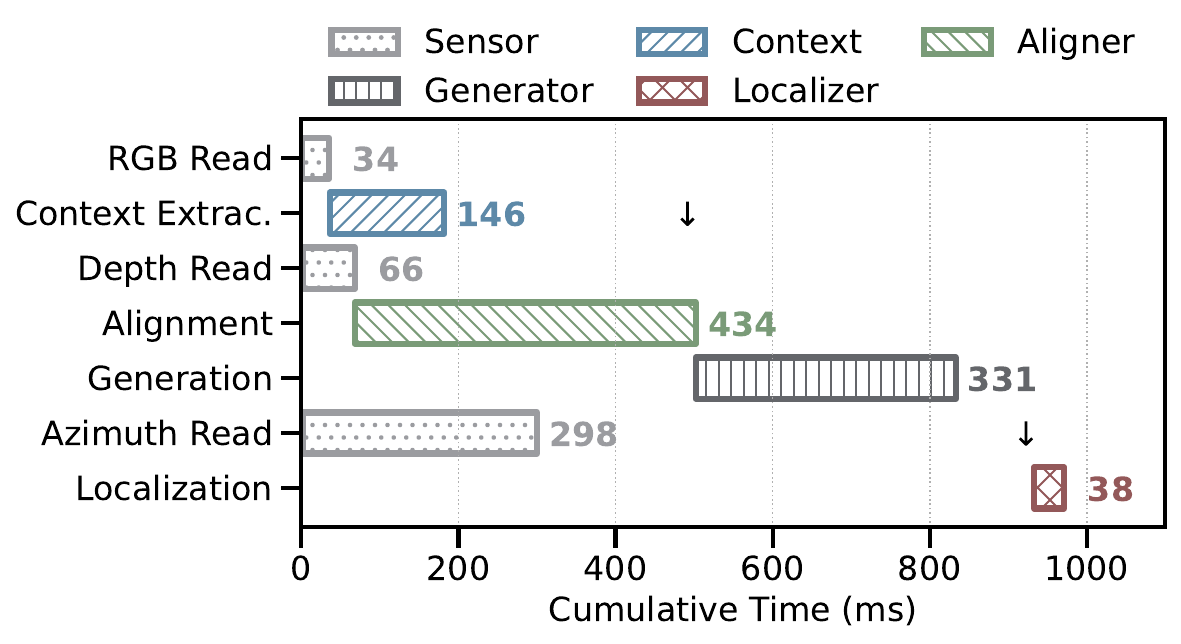}
\vspace{-3mm}
\caption{Response time of \sys}
\label{fig:response_time}
\vspace{-3mm}
\end{figure}

\section{Discussion}

\parlabel{Why and When Visual Context Helps.}
\bnote{\sys leverages visual context (RGB-D) to capture scene geometry and surface characteristics that influence mmWave propagation. Radar-only methods often misinterpret benign variations, caused by walls, clothing, or nearby structures, as anomalies. \sys addresses this by synthesizing a context-conditioned baseline radar spectrum and identifying anomalies as residual deviations from the actual measurements. This mechanism is particularly beneficial in scenarios, such as through-wall or concealed-object settings, where material attenuation and environmental clutter significantly alter radar responses.}

\parlabel{Module Benefits in Through-Wall Scenarios.}
\bnote{Within the \sys pipeline, the Context Extractor identifies wall materials and structural layout to calibrate radar expectations, since different materials (e.g., styrofoam vs.\ particle board) produce distinct attenuation and reflection effects. The Generator then synthesizes a context-conditioned ``blank-scene'' radar baseline that models expected propagation without human presence. Unlike radar-only change detection methods that compare consecutive frames ($t$ vs.\ $t\!-\!1$) and rely on motion cues, \sys\ compares observed signals against a synthesized context-aware baseline, enabling detection of stationary intruders or fallen subjects. Rather than replacing task-specific signal processing approaches such as RTWLBR~\cite{wang2023realtime}, \sys\ introduces a cross-modal, residual-based sensing paradigm that generalizes across through-wall and related applications.}

%% file: tex/08_literature.tex
\section{Related Work}

\parlabel{Deep Anomaly Detection.}
Autoencoders (AEs), Variational Autoencoders (VAEs), and Generative Adversarial Networks (GANs) are widely used for deep anomaly detection. AEs reconstruct inputs and use reconstruction error for anomaly scoring~\cite{pang2021deep}, but tend to overfit and reconstruct outliers. VAEs improve generalization by enforcing a latent prior~\cite{wang2024design}. GANs detect anomalies through poor reconstructions or low discriminator confidence~\cite{liu2023simplenet, zhuang2024research}, but are unstable and sensitive to hyperparameters. While effective for visual data, these models assume unimodal inputs and struggle with radar signals that exhibit complex physical artifacts. Our work introduces a cross-modal generator conditioned on visual cues to improve anomaly detection in mmWave domains.

\parlabel{mmWave Anomaly Detection.}
Deep learning has been applied to mmWave anomaly detection in autonomous driving, wireless sensing, and security. \textit{GANomaly}~\cite{akcay2019ganomaly} detects radar-based anomalies but lacks spatial grounding. In wireless domains, CGANs and VAEs are used for spectral anomaly detection~\cite{toma2020deep}, but focus on 1D features and overlook spatial context. \textit{MMW-Carry}~\cite{gao2024mmwcarry} uses mmWave-RGB fusion for object classification but depends on labeled data and cannot detect novel anomalies. Other works~\cite{yang2021realtime} apply GANs for fusion or enhancement, not anomaly modeling. In contrast, our method enables localized, context-aware anomaly detection using generative modeling and visual semantics, improving generalization to diverse and unseen cases.

\parlabel{mmWave Applications.}
Recent mmWave systems enable high-resolution human monitoring, but most are task-specific and lack open-set anomaly reasoning. For fall detection and behavior monitoring, prior works such as Shen et al.~\cite{shen2024advanced}, LT-Fall~\cite{zhang2023ltfall}, Zhao et al.~\cite{zhao2025mmfall}, and Li et al.~\cite{li2022realtime} demonstrate effective tracking or fall recognition but rely on predefined motion patterns and do not generalize to unseen anomalies. Through-wall human localization methods, including Wang et al.~\cite{wang2023realtime}, focus on multi-human tracking behind obstructions but remain closed-set and non-generative. Other systems—\textit{mmEye}~\cite{zhang2021mmeye}, \textit{Argus}~\cite{duan2025argus}, \textit{mmFace}~\cite{xu2022mmface}, \textit{mmSpyVR}~\cite{mei2024mmspyvr}, and \textit{Metasight}~\cite{woodford2024metasight}—achieve high-resolution imaging or authentication through materials, yet they operate under supervised conditions and cannot reason about unknown anomalies. In contrast, our work enables spatial anomaly localization under occlusion by combining generative modeling with visual context, supporting open-set anomaly detection across diverse mmWave environments.

%% file: tex/09_conclusion.tex
\section{Conclusion}
This paper presents \sys, a multi-modal anomaly detection framework that integrates mmWave radar and visual inputs to detect and localize signal-level anomalies in occluded or visually inaccessible scenarios. By aligning RGB-depth data to radar perspective, extracting contextual cues, and synthesizing expected mmWave spectra through cross-modal diffusion, \sys enables fine-grained anomaly localization using a dual-branch transformer. Experimental results across diverse applications, including through-cloth weapon detection, through-wall intrusion detection, and through-wall fall detection, demonstrate that \sys consistently outperforms prior methods in both accuracy and localization precision. The system's ability to incorporate visual context enhances robustness to clothing materials, environmental clutter, and structural barriers. Overall, \sys advances the state of mmWave anomaly detection by combining generative modeling and context-awareness, paving the way for broader deployment in real-world security, safety, and monitoring applications.
